\documentclass[authoryear,final,5p,times,twocolumn]{elsarticle}
\usepackage{amssymb}
\usepackage{amsmath}
\usepackage{xcolor}
\usepackage{xurl}
\usepackage{booktabs}
\emergencystretch=\maxdimen
\usepackage[colorlinks = true,
            urlcolor  = black]{hyperref}
            
\usepackage{graphicx}
\usepackage{multirow}
\usepackage{caption}
\usepackage{stfloats}
\usepackage{bbm}
\journal{}
\journal{ISPRS Journal of Photogrammetry and Remote Sensing}

\begin{document}

\begin{frontmatter}
\title{\textsc{DeepC4}: Deep Conditional Census-Constrained Clustering\\for Large-scale Multitask Spatial Disaggregation of Urban Morphology}
\author[label1,label3]{Joshua Dimasaka\corref{cor1}}
\author[label4,label5]{Christian Gei{\ss}}
\author[label1,label3]{Emily So}
\affiliation[label1]{
            organization={Department of Architecture, 
            University of Cambridge},
            city={Cambridge},
            country={United Kingdom}}
\affiliation[label3]{
            organization={Cambridge University Centre for Risk in the Built Environment},
            city={Cambridge},
            country={United Kingdom}}
\affiliation[label4]{
            organization={Earth Observation Center, 
            German Aerospace Center},
            city={We{\ss}ling},
            country={Germany}}
\affiliation[label5]{
            organization={Institute of Geography,
            University of Bonn},
            city={Bonn},
            country={Germany}}
\cortext[cor1]{Corresponding Author: jtd33@cam.ac.uk}

\begin{abstract}
To understand our global progress for sustainable development and disaster risk reduction in many developing economies, two recent major initiatives – the Uniform w Exposure Dataset of the Global Earthquake Model (GEM) Foundation and the Modelling Exposure through Earth Observation Routines (METEOR) Project – implemented classical spatial disaggregation techniques to generate large-scale mapping of urban morphology using the information from various satellite imagery and its derivatives, geospatial datasets of the built environment, and subnational census statistics. However, the local discrepancy with well-validated census statistics and the propagated model uncertainties remain a challenge in such coarse-to-fine-grained mapping problems. Therefore, we present \underline{D}eep \underline{C}onditional \underline{C}ensus-\underline{C}onstrained \underline{C}lustering (\textsc{DeepC4}), a novel deep learning-based spatial disaggregation approach that incorporates local census statistics as cluster-level constraints while considering multiple conditional label relationships in a joint multitask learning of the patterns of satellite imagery. As a demonstration using Rwandan urban morphology, \textsc{DeepC4} achieves macro-F1 scores of 0.63, 0.78, and 0.45 and macro-mIoU of 0.57, 0.71, and 0.42 for roof, wall, and height prediction respectively, estimates national dwelling and occupant counts within 1.13\% and 1.11\% error compared to census records, outperforming GEM (2.03\% and 3.29\%), and occupies 32\%-49\% more 500-meter grid pixels than METEOR across provinces, reflecting improved spatial coverage consistent with 2022 urban development. As the world approaches the conclusion of many global frameworks in 2030, our work offers a new deep learning-based mapping technique that explicitly encodes well-validated census and experts' belief systems to achieve an explainable and interpretable auditing of existing coarse-grained derived information at large scales.
\end{abstract}


\end{frontmatter}

\newpage
\section{Introduction}
As one of the key priorities of the 2015-2030 Sendai Framework for Disaster Risk Reduction \citep{un2015sfdrr} alongside the targets of the 2030 Agenda for Sustainable Development \citep{un2015sdg}, the need to understand disaster risk at large scales has urged recent various multi-country projects such as the METEOR project for UN-designated Least Developed Countries (LDCs) \citep{huyck2019meteor} and the Global Earthquake Model (GEM) Foundation Uniform Africa Exposure Data \citep{paul2022development} to map the locations of populated human settlements and buildings in many developing economies. Assessing disaster risk complements this information on the exposed location with the characterization of physical vulnerability, which is commonly described in terms of the type of construction material and the height of a building. For example, the standard taxonomies of GEM \citep{brzev2013gem}, PAGES-STR \citep{jaiswal2010global}, HAZUS \citep{kircher2006hazus}, and EMS-98 \citep{grunthal1998ems} provide a classification system on building typologies, indicating the inherent structural capacity or vulnerability of a building. More broadly, in a regional scale, this characterization of the physical attributes of buildings collectively contributes to the three-dimensional urban form or morphology of a place \citep{labetski20233d}.

However, while such large-scale efforts using classical spatial disaggregation techniques have revealed a comprehensive understanding of the urban morphology indicators used for regional disaster risk assessment, its numerical discrepancy with well-validated census records at the local level has remained a significant challenge in establishing a general consensus between the data-driven mapping techniques and the local knowledge and practice \citep{wardrop2018spatially, petrarulo2022meteor}, thereby limiting its use by the wider community such as urban planners and disaster risk managers on the ground \citep{gevaert2021fairness}. Although other finer-grained efforts, such as building-level attribute identification, have shown reliable predictive performances \citep{geiss2018cost, gouveia2024automated, silva2024building}, these alternatives are data-expensive, resource-intensive, and scope-limited, when scaling up to a larger areal extent. Thus, the tradeoff between the resolution quality of mapping and the accuracy of information has become a primary issue in supporting the evidence-based decisions of many global frameworks \citep{clark2023investigating}.

Despite the potential wide-scale offerings of previous studies, their use of secondary products and input satellite imagery with varying temporal signatures further compounded and propagated the spatial uncertainties of the resulting maps of urban morphology indicators. For example, even though both GEM and METEOR used the rich information from remote sensing data and the data-driven techniques such as the random forest tool \citep{stevens2015disaggregating} for WorldPop population map \citep{tatem2017worldpop} and the symbolic machine learning \citep{blei2018accuracy} for Global Human Settlement Layer \citep{pesaresi2024advances}, the performance of WorldPop for Kenya still achieved 60\% \citep{stevens2015disaggregating}. In addition, the METEOR project used varying temporal information of their input geospatial maps due to the unavailability of temporally coherent data during the time of analysis \citep{huyck2019meteor}. As a result, these challenges inherent to the model uncertainties still necessitate subsequent careful validation and other independent analyses for suitability and overall quality \citep{gevaert2021fairness}. 

Nevertheless, the increasing availability of public EO data and acquisition platforms such as Google Earth Engine \citep{gorelick2017google} has recently enabled many new opportunities and improvements in various geospatial modelling of direct and indirect indicators of urban morphology such as the spatial disaggregation of building energy consumption \citep{zhou2023high}, population \citep{darin2022population}, building material type using height distribution \citep{geiss2023benefits}, and settlement structure characteristics \citep{meinel2009analyzing}. While such region-specific examples have demonstrated the simplicity and flexibility in dealing with the non-uniform, highly diverse, and heterogeneous classifications at large scales, the quality of the coarse-to-fine-grained disaggregation task remains weakly supervised. For example, a particular building height category can correspond to multiple possible classifications of wall and roof material, thereby introducing multiple one-to-many, many-to-many, or fuzzy combinations \citep{pittore2018risk, gomez2022epistemic}. Specifically, a residential wooden building has 40-75\%, 25-40\%, and 0-20\% chances to have one, two, and three storeys, respectively, according to the outcomes of the SSAHARA survey of the expert belief system on the local indigenous knowledge of construction practice \citep{paul2022development}. Because of the limited generalizability due to the weak and conditional supervision, numerous studies have similarly implemented deep constrained clustering techniques in renewable electricity modelling \citep{seljom2021stochastic} and agricultural crop management \citep{lampert2019constrained, lafabregue2025samarah} to extend the model capability on high-dimensionality and non-linearity of EO data, with the consideration of coarse-grained label information as cluster-level constraints that can provide performance guarantee \citep{bradley2000constrained, wagstaff2001constrained, demiriz2008using, zhu2010data}.

Therefore, we develop the Deep Conditional Census Constrained Clustering (\textsc{DeepC4}), a novel deep learning-based spatial disaggregation approach that fundamentally addresses the numerical discrepancy with coarse-grained or cluster-level labels (e.g., census statistics) and incorporates conditionality on multiple possible labels (i.e., from experts' belief system gained from local indigenous knowledge) in a multitask optimization, while still exploiting the richness of information that can be derived from near-real-time publicly available satellite imagery and other derived products. This coarse-to-fine-grained approach can also be generalizable to other weakly supervised problems that involve the use of EO data for urban informatics applications with encoded constraints from other external validation surveys. By considering the explicit importance of well-validated census and experts' belief systems, our approach also contributes to the overall consistency of resulting deep learning-based maps, thereby improving their relative explainability and interpretability with respect to the current understanding of end users. In this study, we demonstrate the application of \textsc{DeepC4} using the case of Rwanda, one of the LDCs that has experienced a significant technological difficulty in advancing exposure mapping and its physical vulnerability characteristics. 

The remainder of this paper is organized as follows: Section \ref{Materials} details the information about census statistics, expert belief systems on the conditional relationships of urban morphology, satellite imagery, and built-area datasets. Section \ref{Methods} describes the preprocessing steps, the iterative construction of a probable space of building presence for disaggregation implementation, and the \textsc{DeepC4} algorithms on dimensional reduction and joint multitask learning optimization. Section \ref{ResultsAndDiscussion} compares our findings with the corresponding outputs of the METEOR project and the GEM for the entire country of Rwanda, and explains the numerical evaluations of the performance of \textsc{DeepC4} in achieving learning stability while minimizing the errors in the clustering implementation.  

\section{Materials}
\label{Materials}

\subsection{Study Area}
\begin{figure}[b!]
    \centering
    \includegraphics[width=80mm]{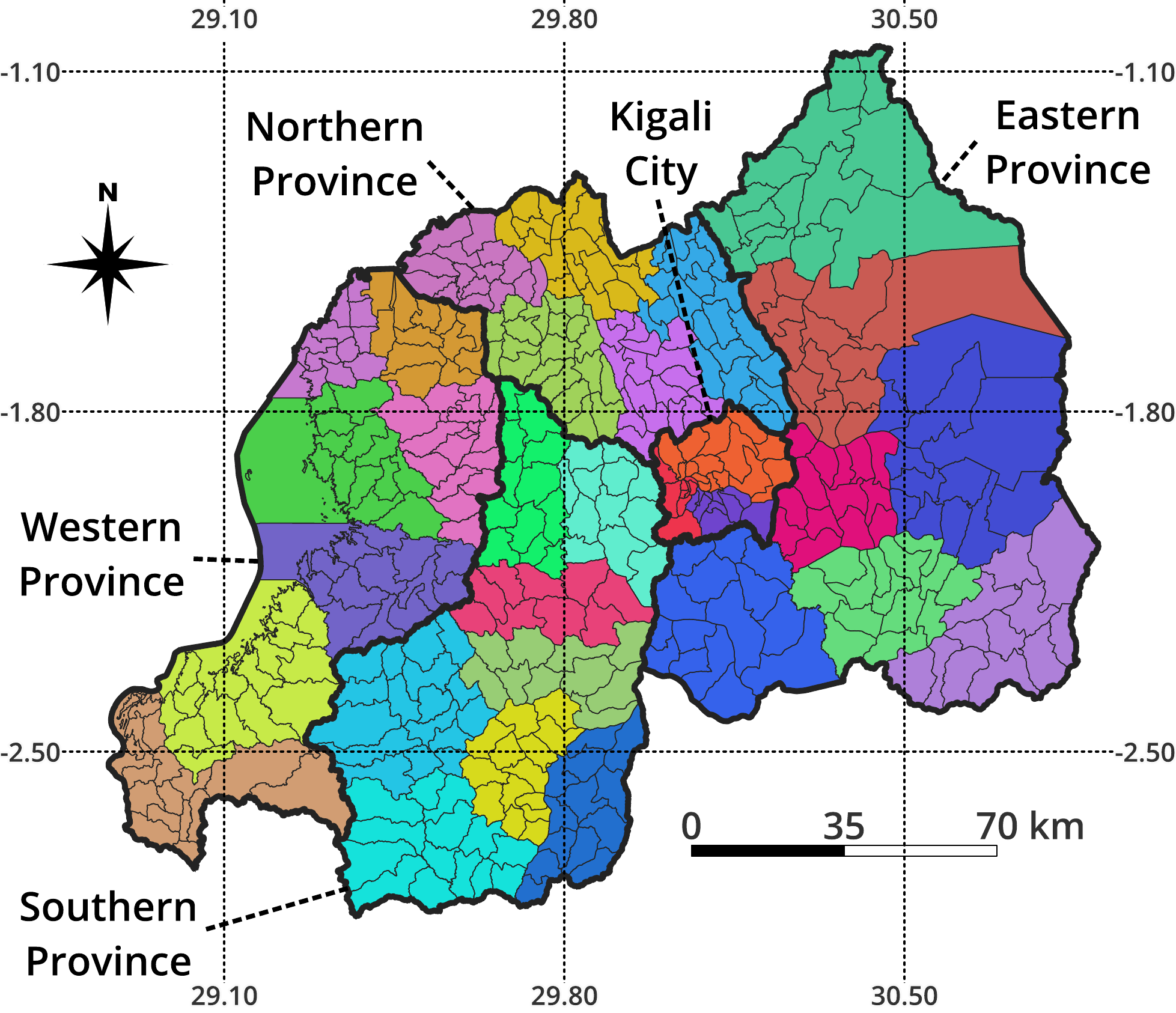}
    \caption{Geographical extent of Rwanda \citep{gadm}. The boundaries of sectors and provinces are respectively displayed with thinner and thicker lines. Every distinct color represents a district.}
    \label{studyArea}
\end{figure}

With over 26,300 square kilometers and located in the east-central region of the African continent, the densely populated country of Rwanda consists of 416 sectors, 30 districts, and 5 provinces, as shown in \autoref{studyArea}.

\subsection{Census Statistics}
From the Fifth Population and Housing Census (RPHC5) 2022, with acceptable data quality at a net coverage rate of 98.7\% and a rate of erroneous inclusion of 0.2\% \citep{nisr}, we used the sector-level census statistics as cluster-level constraints to \textsc{DeepC4}. Specifically, we obtained 416 rows of data on average household size, urban and rural population counts, 14 wall material classes, and 9 roof material classes. 

For the year 2022, the average household size ranged from 3.05 to the highest value of 5.46 population per household unit that was achieved by Nkombo, Rusizi, Western Province. The least and most populated sectors were Gitovu, Burera, Northern Province at 11,531, and Kinyinya, Gasabo, Kigali City at 125,400, respectively. Across all sectors, the most frequent wall and roof materials were sun-dried bricks and iron sheets, respectively.

At the national scale, \citep{paul2022development} reported that only 1.4\% and 0.6\% respectively correspond to the commercial and industrial occupancies of the 2020 building counts, whereas the majority at 98\% is residential. Hence, we further assumed that our study is primarily concerned with the number of residential buildings with minimal underestimation effect on the overall results at the sectoral level.

\begin{table*}[tp]
    \centering
    \captionsetup{justification=centering}
    \caption{Urban morphology indicators and their respective classes and descriptions in the resulting Rwandan maps.}
    \label{table:classes}
    \resizebox{\textwidth}{!}{%
    \begin{tabular}{lcl}
    \hline
    \textbf{Indicator} &
      \multicolumn{1}{l}{\textbf{Number}} &
      \textbf{Classes} \\ \hline
    Roof &
      4 &
      Iron Sheets; Local, Industrial, and Asbestos Tiles; Concrete; Grass \\ \hline
    Wall &
      8 &
      Wood with mud; Sun-dried bricks; Cement bricks; Burnt bricks; Stones; Concrete; Timber; Others \\ \hline
    Height &
      6 &
      \begin{tabular}[c]{@{}l@{}}H:1 (1 storey); H:2 (2 storeys); H:3 (3 storeys)\\ HBET:3-6 (3-6 storeys); HBET:4-6 (4-6 storeys); HBET:8+ (8 or more storeys)\end{tabular} \\ \hline
    \begin{tabular}[c]{@{}l@{}}Macro\\ Taxonomy\end{tabular} &
      16 &
      \begin{tabular}[c]{@{}l@{}}
      CR/LFINF (reinforced concrete with infill walls)\\ 
      CR/LWAL (reinforced concrete with shear walls)\\ 
      MATO (others such as reeds, plastics, and fabrics)\\
      MCF+CB/LWAL (confined concrete block masonry with shear walls)\\ 
      MCF+CL/LWAL (confined clay brick masonry with shear walls)\\ 
      MUR+ADO+MOC/LWAL (unreinforced adobe block masonry with cement mortar and shear walls)\\ 
      MUR+ADO/LWAL (unreinforced adobe block masonry with shear walls)\\ 
      MUR+CB/LWAL (unreinforced concrete block masonry with shear walls)\\ 
      MUR+CL+MOC/LWAL (unreinforced clay brick masonry with cement mortar and shear walls)\\ 
      MUR+CL/LWAL (unreinforced clay brick masonry with shear walls)\\ 
      MUR+STDRE+MOC/LWAL (unreinforced dressed stone masonry with cement mortar and shear walls)\\ 
      MUR+STDRE/LWAL (unreinforced dressed stone masonry with shear walls)\\ 
      MUR+STRUB+MOC/LWAL (unreinforced rubble stone masonry with cement mortar and shear walls)\\
      MUR+STRUB/LWAL (unreinforced rubble stone masonry with shear walls)\\
      W+WWD/LWAL (wattle and daub with shear walls)\\ 
      W/LWAL (wood with shear walls)
      \end{tabular} \\ \hline
    \end{tabular}%
    }
\end{table*}

\subsection{Experts' Belief System on Construction Practice}

We considered the valuable encoded mapping scheme of \citet{paul2022development}, which creates discrete probabilistic relationships among wall material, height, and macro-taxonomy classes, since it is based on a comprehensive engagement and judgment of local experts on the construction methodology. This information controls the learning behavior of \textsc{DeepC4} using the numerical conditions among these urban morphology indicators. As presented in \ref{appendix:conditionalProb}, this explicit encoding of discrete conditional probability values supports the interpretability of resulting maps since the inferred distribution of classes, including any deviations, can be explained by the probabilistic relationships.

\autoref{table:classes} enumerates the complete list of urban morphology indicators in our resulting Rwandan maps, which were reduced from the original literature \citep{paul2022development} due to the overlapping general characteristics and additional computational benefit in the implementation of constrained clustering (i.e., lower number of categories results in less computational demand). However, it is important to note that, in this study, these presented indicators are limited only to physical building-level attributes, but still contribute to the understanding of the urban morphology of a place (e.g., spatial patterns, density, and other numerical metrics).

\subsection{Earth Observation Data}

We obtained the following pre-processed 10-meter satellite imagery whose signals serve as a proxy to infer the distribution of distinct urban morphology indicators for the 2022 period via Google Earth Engine \citep{gorelick2017google}. The contribution of each input modality to predictive performance is empirically evaluated in the ablation study in \autoref{sec:ablation}.

\textbf{Sentinel-1 SAR GRD.} Using its cloud-penetrating capability for surface roughness sensitivity \citep{koppel2017}, we used the mean of the Ground Range Detected (GRD) scenes acquired from the dual-polarization C-band Synthetic Aperture Radar (SAR) instrument at 5.405 GHz of the Sentinel-1 satellite \citep{sentinel1}. It consists of two bands: VV (vertical transmit, vertical receive) and VH (vertical transmit, horizontal receive) signals. To avoid data incompleteness across large areas, we disregarded filtering by orbital number and satellite direction.

\textbf{Sentinel-2 Harmonized MSI.} Using its multispectral imaging capability for optical signatures \citep{braun2019}, we also extracted the median of the atmospherically corrected surface reflectance signals represented by 10 bands -- red (B4), green (B3), blue (B2), red edge 1 (B5), red edge 2 (B6), red edge 3 (B7), red edge 4 (B8A), near infrared (B8), short-wave infrared 1 (B11), and short-wave infrared 2 (B12) -- that are acquired from the MultiSpectral Instrument (MSI) of the Sentinel-2 satellite \citep{sentinel2}. The aggregation by year also enables minimizing the unnecessary noisy cloudy or shadowy signals using the available and corresponding Sentinel-2 cloud probability dataset \citep{sentinel2cloud}.

\subsection{Built Area Information}
\label{sec:builtareainfo}

Given a large areal extent with several land cover types, information about the locations of built areas, such as building footprints, refines the number of potential locations to be assigned with the previously mentioned urban morphology indicators. Despite the growing efforts in mapping building footprints for multi-year periods, temporally disaggregated building footprints by year remain scarce due to the limited and expensive year-specific groundtruth labels. Thus, we performed an iterative construction of possible space of disaggregated locations of building presence for the year 2022 using a combination of various imperfect building footprint datasets -- OpenStreetMap \citep{geofabrikOSM}, Google Open Buildings V3 \citep{sirko2021continental}, Microsoft Building Footprints \citep{microsoft}, and Overture Foundation Building Footprints \citep{overture} -- and the 2022 map of average probability of complete coverage by built area from the Dynamic World V1 dataset, which is a near-real-time land cover map derived from Sentinel-2 satellite imagery \citep{brown2022dynamic}. 

In our study, while it may be desirable to develop our own data-driven model to independently extract 2022 building footprints directly from EO data in our problem setting, we note that this entails the acquisition of reliable building groundtruth labels for the particular year. In contrast, because of its limited availability, we instead relied on the multi-year building footprints from these various sources and iteratively constructed a set of probable map locations (i.e., pixels) with the aid of the Dynamic World V1 dataset, which has a higher temporal resolution that enables a reliable representation for the year 2022. This iterative procedure addressing this data limitation is further discussed in Section \ref{subsec:iterativeDisaggregated}. Nevertheless, this approach remains valid for the demonstration of \textsc{DeepC4}, which focuses on the deep representation and constrained clustering.

\subsection{Building-level Groundtruth Information}

While a vanilla unsupervised clustering does not necessitate any groundtruth information, our \textsc{DeepC4}, however, involves the training of deep neural network parameters to learn a more meaningful latent reduced representation of EO satellite imagery signals using building-level groundtruth information, which effectively supervises the constrained clustering. Because such groundtruth information in the form of a 2022 Rwandan building footprint is not widely available and may often use a different modality (e.g., different classification system), we instead used the weak and limited supervision provided by the 2015 dataset on building typology for the Kigali City, Rwanda, which covers 30 (7.2\%) out of all 416 sectors \citep{bachofer2019building}. 

As a limitation, our empirical results assumed that the 2015 groundtruth information has a minimal difference with the 2022 true information by defining a common subset of map locations with valid pairs of the post-processed 2015 groundtruth information and \textsc{DeepC4} 2022 predictions, as presented with further details in Section \ref{subsubsec:jointmultitask}. This implies that any subsequent interpretations are contextually limited by the geographic and temporal biases inherent in using only the available supervising information from Kigali City, Rwanda, thereby potentially disregarding distinct urban-rural differences. Nonetheless, while it may be desirable to have representative samples across the country of Rwanda, our demonstration recognizes that this current data gap reflects an important domain challenge and presents an opportunity for future fine-tuning improvement, when a large-scale survey of Rwandan building typology becomes available.


\section{Methods}
\label{Methods}

\begin{figure*}[h]
    \centering
    \includegraphics[width=7.2in]{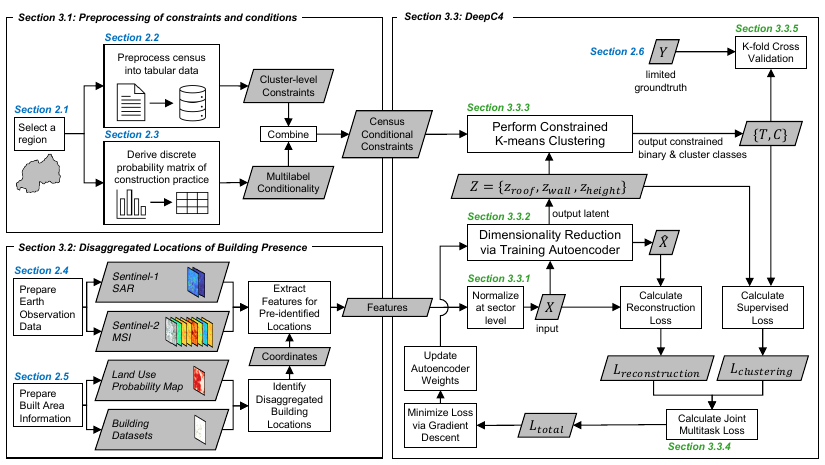}
    \caption{\textsc{DeepC4} Implementation. From left to right, we began with the encoding of census conditional constraints, preparations of EO features, and selection of disaggregated building locations from various imperfect sources. We then trained an autoencoder to obtain a set of reduced-dimension latent representations that were used for constrained clustering algorithms. Jointly trained using the reconstruction and prediction losses, we evaluated the performance of the autoencoder using the available building-level groundtruth.}
    \label{fig:DeepC4}
\end{figure*}

In this section, we present a detailed summary of the major procedures from the preprocessing of census information into conditional constraints and the selection of disaggregated locations of building presence using various available datasets to the implementation of our proposed \textsc{DeepC4}, which consists of five sections on normalization, dimensional reduction, constrained clustering, joint multitask learning, and performance evaluation, as illustrated in \autoref{fig:DeepC4}.

\subsection{Preprocessing of constraints and conditions}

The preprocessing started with the encoding of census statistics from 30 district-level reports into a machine-readable format. For every sector $s_i$, where $i = 1, 2, ..., N_s$ and $N_s$ corresponds to the total number of sectors in the country, we computed its corresponding average household size, $h_{s_i}$,  by taking the ratio of sector-level population and private household count. Because the census also reported the composition of the population by urban and rural areas, we then computed the number of dwellings for urban ($d_{{s_i},urban}$) and rural ($d_{{s_i},rural}$) by respectively dividing the urban and rural population by $h_{s_i}$. 
By respective conditional proportions, we distributed $d_{{s_i},urban}$ and $d_{{s_i},rural}$ into wall, material, height, and macro-taxonomy classes using the probabilistic relationships in \ref{appendix:conditionalProb}. As a result, for both rural and urban areas of sector $s_i$, we obtained the number of buildings for each urban morphology indicator in \autoref{table:classes}. 

However, due to the lack of temporally consistent data on building footprints in a vectorized or polygonal format, as previously discussed in Section \ref{sec:builtareainfo}, we transformed the building counts by approximating their equivalents to the rasterized, image, or pixel-based representation. We note that this approach may be sensitive to the vector-to-raster conversion because of the differences between the grid configuration and the orientation and geometry of the enclosed building footprints. 

Thus, to enable a coherent rasterized analysis, by using a hard assumption wherein a building has a 60-m\textsuperscript{2} ground floor area similar to \citet{paul2022development}, we converted the number of buildings into the number of grid pixels at 10-m scale, which is the representative spatial resolution of our EO satellite imagery signals. This follows that a 10-m grid pixel on the map, whose area is 100 m\textsuperscript{2}, contains approximately 1.67 units of 60-m\textsuperscript{2} buildings (i.e., 100 m\textsuperscript{2} per grid / 60 m\textsuperscript{2} per building). We used the resulting number of grid pixels as our transformed constraints in the succeeding section. When high-resolution imagery becomes available, it follows that this approximating assumption can remain valid, with further reduced uncertainty brought about by the sensitivity to vector-to-raster conversion.

\subsection{Disaggregated locations of building presence}
\label{subsec:iterativeDisaggregated}

Due to the imperfect nature of our built area information, for every sector $s_i$, we iteratively constructed a set of possible grid pixels that could represent building presence. The overall procedure mainly relies on determining a threshold between 0 and 1 that interacts with the 2022 map of average probability of complete coverage by building area from the Dynamic World V1 dataset \citep{brown2022dynamic} until the number of grid pixels, ($n_{constraint}$), at a particular threshold matches the transformed cluster-level constraints (i.e., the required number of grid pixels).

Using the rasterized footprints from the combined four building datasets within the geographical extent of sector $s_i$, we first obtained the number of grid pixels with rasterized building footprint signals ($n_{bldg}$) and then evaluated its difference from the expected number of buildings that was previously calculated from the transformed constraints. If $n_{bldg}$ did not exceed $n_{constraint}$, we obtained additional grid pixels to match $n_{constraint}$. Using the 2022 map of average probability of complete coverage by building area, we gradually decreased the probability threshold starting from the largest probability associated with $n_{bldg}$ until the difference became negligible. Similarly, if $n_{bldg}$ was more than $n_{constraint}$, we removed excess grid pixels to match $n_{constraint}$ by gradually increasing the probability threshold starting from the smallest probability associated with $n_{bldg}$ until the difference also became negligible. Using the calculated probability threshold, we prepared the final set of valid grid pixels with $M$ distinct locations that \textsc{DeepC4} used to construct the feature array of EO satellite imagery signals.

\subsection{DeepC4}

Unlike the two previous sections, which are applied to all 416 sectors, the succeeding sections on the implementation of \textsc{DeepC4} relied on only 20 sectors with available building-level groundtruth information, which are all primarily located within and near the most populous province of Kigali City. 

\subsubsection{Sector-level normalization}

For every sector $s_i$ of the 20 sectors, to achieve a properly scaled feature array as input to our \textsc{DeepC4}, we normalized the logarithm of each EO satellite imagery band (e.g., red or B4) across all $M_{i}$ distinct locations with zero mean and one standard deviation. For location encoding, we also normalized the row and column indices of its grid pixels. Thus, we prepared a feature array, $X_{s_i}\in\mathbb{R}^{\left|M_{i}\right|\times 14}$,  with varying first dimension, which depends on the size of $M_{i}$ distinct locations, and a fixed 14 columns corresponding to the number of EO satellite imagery signals including the two location encoding features. This resulted in a combination of 127,757 grid pixels across all 20 sectors.

\subsubsection{Dimensional reduction}

For computational efficiency and consistency of constrained clustering implementation, we extracted a low-dimension latent representation, $Z_{s_i}$, from the high dimensionality of $X_{s_i}$ using an autoencoder, a deep non-linear reduction technique. While the use of other image-based deep learning techniques may be desirable to capture spatial or positional information, these approaches require an approximate transformation of cluster-level constraints into corresponding image tiles. The autoencoder addressed this limitation by incorporating $X_{s_i}$ with location encoding features in a tabular format, as the cluster-level constraints are only expressed at the sector level and not in a form suitable for image tile representation.

Following a fully connected layer network architecture, our study assigned an arbitrary number of latent channels to each urban morphology indicator, as shown in \autoref{fig:DeepC4}. In this study, we used a total of three latent channels for roof ($z_{s_{i},roof}$), wall ($z_{s_{i},wall}$), and height ($z_{s_{i},height}$) in such a way that each separate latent channel is correspondingly interpretable to individual urban morphology indicator. Across all 20 sectors, we trained a single set of deep neural network parameters until the model performance converged, indicating that the learned latent representation has achieved a desirable clustering behavior given the conditional constraints under consideration. 

Using the adaptive moment estimation (Adam) algorithm \citep{kingma2014adam} at a learning rate of 0.0001, a square gradient decay factor of 0.95, and a gradient decay factor of 0.80, we evaluated the autoencoder reconstruction loss in terms of mean-squared error (MSE), which is calculated for every sector $s_{i}$ (see \autoref{eq:reconstruction}), and combined it for joint optimization with the calculated loss from the constrained clustering in the succeeding section.

\begin{equation}
    \label{eq:reconstruction}
    L_{s_{i},reconstruction} = \frac{1}{|M_{i}|}\sum_{j=1}^{|M_{i}|}(x_j-\hat{x_j})^2
\end{equation}

\noindent where $x_j$ is the $j^{th}$ element of the sector-level normalized feature array $X_{s_i}$ with size $|M_{i}|$ grid pixels and $\hat{x_j}$ is its corresponding reconstruction.

\subsubsection{Constrained clustering}

Prior to the clustering implementation, we processed the building-level groundtruth information because the building typology classification of \citet{bachofer2019building} (e.g., building height as a continuous variable) is different from the classes of every urban morphology indicator presented in \autoref{table:classes} (e.g., building height as a discrete and overlapping variable). Upon inspection, we created a mapping scheme to match their respective shared and overlapping characteristics, as presented in \ref{appendix:mappingscheme}. Similar to the fuzzy scoring approach of \citet{pittore2018risk}, such assumptions are necessary, particularly in the limited and weak supervision setting of this domain problem, to demonstrate the implementation of \textsc{DeepC4}, thereby generalizing for cases with more precise building-level groundtruth information as it becomes available.

The \textsc{DeepC4} extended the constrained k-means clustering algorithm developed by \citet{bradley2000constrained} with the low-dimension latent representation learning to exploit the rich information of EO data using deep and flexible neural networks and the joint multitask learning across three urban morphology indicators, resulting in a k-means-friendly space with optimal and stable clustering behavior that is weakly supervised by conditional constraints and available building groundtruth information.

Following the mathematical formulation of \citet{bradley2000constrained} and modifying it into the context of our study, \textsc{DeepC4} implements that, for every sector $s_{i}$ with corresponding latent representation $Z_{s_i} = \left\{ z^{j}_{s_{i},roof}, z^{j}_{s_{i},wall}, z^{j}_{s_{i},height}  \right\}_{j=1}^{|M_i|}$ with size of $|M_i|$ grid pixels, we determined the optimal $k$ target cluster centers $C^{1}_{s_i}, C^{2}_{s_i}, \ldots, C^{k}_{s_i}$ with minimized sum of the 2-norm distance squared between each $j^{th}$ point of $Z_{s_i}$ and its set of nearest cluster center $C^{h}_{s_i} = \left\{ c^{h}_{s_{i},roof}, c^{h}_{s_{i},wall}, c^{h}_{s_{i},height}  \right\}_{h=1}^{k}$. In symbols,

\begin{equation}
    \label{eq:minmin}
    \tag{2}
    \min _{C^{1}_{s_i}, \ldots, C^{k}_{s_i}} \sum_{j=1}^{|M_i|} \min _{h=1, \ldots, k}\left(\frac{1}{2}\left\|Z_{s_i}-C^{h}_{s_i}\right\|_{2}^{2}\right)
\end{equation}

Reframing \autoref{eq:minmin} with the use of a binary selection array $T_{j, h}\in\mathbb{R}^{\left|M_{i}\right|\times k}$ (i.e., $T_{j, h}=1$ if, for example, the latent representation $z^{j}_{s_{i},roof}$ is closest to center $c^{h}_{s_{i},roof}$ and zero otherwise) gives:
\[
\begin{array}{lc}
\underset{C, T}{\operatorname{minimize}} & \sum_{j=1}^{|M_i|} \sum_{h=1}^{k} T_{j, h} \cdot\left(\frac{1}{2}\left\|Z_{s_i}-C^{h}_{s_i}\right\|_{2}^{2}\right) \\
\text { subject to } & \sum_{h=1}^{k} T_{j, h}=1, j=1, \ldots, |M_i|  \tag{3}\\
& T_{j, h} \in \left\{0,1\right\}, j=1, \ldots, m, h=1, \ldots, k
\end{array}
\]

\noindent Iteratively solving for $T_{j, h}$ given fixed $C^{h}_{s_i}$ (i.e., cluster assignment) followed by computing the $C^{h}_{s_i}$ given fixed $T_{j, h}$ (i.e., cluster update) achieves the optimal values for $T_{j, h}$ until the changes in $C^{h}_{s_i}$ become negligible. 

The \textsc{DeepC4} considered the conditional constraints as input to the minimum cost network flow optimization algorithm \citep{bradley2000constrained} via integer linear programming \citep{huangfu2018parallelizing} as an equivalent formulation. For example, if each of 1,000 grid pixels has a directed connection to every possible cluster assignment (e.g., $N$ number of roof classes), resulting in an exhaustive network with $N \times 1000$ candidate connections. In this way, the `flow' signifies a selected cluster to which a particular grid pixel will be assigned. Assuming a uniform cost to trigger `flow' or `cluster selection' for every connection between any grid pixel and the chosen cluster, our objective is to achieve a minimum cost in such a way that the total 'flow' to each possible cluster should follow the provided conditional constraints in the beginning because the total 'flow' refers to the desired minimum number of grid pixels for each possible cluster assignment, as presented in \autoref{table:classes}.

Using \autoref{eq:clusteringloss}, we computed a Euclidean distance-based supervised loss using the final assignments for cluster $C^{h}_{s_i}$, learned latent representation, and building-level groundtruth information for each urban morphology indicator.

\begin{equation}
    \label{eq:clusteringloss}
    \tag{4}
    L_{s_{i},clustering} = w_{label} \sum_{j=1}^{|M_i|}  \left[ T_{j, h} \sqrt{(Z_{s_i}-C^{h}_{s_i})^2} \right]
\end{equation}

\noindent where $w_{label}$ is a vector of weighting factors, proportionally accounting for the imbalanced number of overlapping cluster assignments for each label, which sums up to unity. 

\subsubsection{Joint multitask learning}
\label{subsubsec:jointmultitask}

It is important to note that, while the clustering is separately convex-optimizing to find its centers, it still depends on the learned latent representations as a direct result of the training of our autoencoder neural network. This joint minimization of the reconstruction loss (\autoref{eq:reconstruction}) and the clustering loss (\autoref{eq:clusteringloss}), influencing a number of latent representations (i.e., of multiple urban morphology indicators), effectively results in a joint multitask learning. The total loss can be generally expressed as a weighted combination of clustering and reconstruction objectives. In this study, we adopted the unweighted formulation ($\lambda = 1.0$) because our initial training implementation suggested comparable orders of loss magnitude while considering the differences in the discreteness of latent labels and the continuous nature of reconstructed signals. To verify that this design choice does not affect model performance, a sensitivity analysis of alternative loss weightings is reported in \autoref{sec:lambda}.
\begin{equation}
    \label{eq:totalloss}
    \tag{5}
    L_{s_i,\mathrm{total}}
    =
    \lambda\,L_{s_i,\mathrm{clustering}}
    +
    L_{s_i,\mathrm{reconstruction}}
\end{equation}

\subsubsection{Performance evaluation}
\label{subsubsec:jointmultitask}

While clustering is an unsupervised approach, the use of deep representation learning in the form of autoencoder neural networks to infer cluster assignments still requires validation to evaluate the resulting trained autoencoder. Due to the limited sample of 20 sectors only, we performed a five-fold cross-validation wherein we set four sectors for each fold (i.e., 20 sectors $\div$ 5 folds). For five times, we tested and averaged the performance of every model trained using only 16 remaining sectors on each unseen fold with six randomly selected sectors.

In addition, the overlapping characteristics of labels (i.e., many-to-many mapping) between \citet{bachofer2019building} ($Y_{C^{h}}$) and the target classes ($C^{h}_{s_i}$) present a difficulty in using standard accuracy metrics, accounting for shared characteristics of the classes of each urban morphology indicator. The task of assigning binary classification metrics has become complicated (i.e., ambiguous or ill-defined negatives) due to the weak and limited supervision of the available building groundtruth information in a multiclass context. Nonetheless, the validity of predictions can still be conservatively verified against the three derived mapping tables in \ref{appendix:mappingscheme}. Thus, we reported the performance of our resulting trained \textsc{DeepC4} model using the proportion of valid predictions (\autoref{eq:tpr}) with multiple $OR$ operators, which measures the fraction of predictions that satisfy these limitations of label ambiguity. This simplifies the performance evaluation of whether the trained \textsc{DeepC4} model results in a valid (i.e., with a check symbol in the mapping table) or invalid (i.e., otherwise) prediction. 
\begin{equation}
    \label{eq:tpr}
    \tag{6}
    p_{valid} = \frac{\sum X }{|M_i|}
\end{equation}

\noindent where $|M_i|$ indicates the number of grid pixels and:
\begin{equation}
  X=\left\{
  \begin{array}{@{}ll@{}}
    1, & \text{if}\ C^{h}_{s_i} \in Y_{C^{h}} \\
    0, & \text{otherwise}
  \end{array}\right.
  \notag
\end{equation} 

Equally important to the selection of the final trained model is a modified $p_{valid}$, which accounts for the imbalanced number of target classes. For each sector $s_i$ with $k$ target classes, we used \autoref{eq:tprmod} to compute a vector of corresponding weight factors, $r^{h}_{s_i}$ to be multiplied to each $p_{valid}$ from \autoref{eq:tpr}.

\begin{equation}
    \label{eq:tprmod}
    \tag{7}
    r^{h}_{s_i} = \frac{|M_i|}{k \times |C^{h}_{s_i}|} 
\end{equation}

\noindent where $|C^{h}_{s_i}|$ is the number of grid pixels under a given target class or cluster $h$.

To enable standard benchmarking, we further adapted per-class classification metrics to the validity-based formulation. Let $P_{j,h} \in \{0,1\}$ denote the prediction matrix, where $P_{j,h} = 1$ indicates that grid pixel $j$ is assigned to class $h$, and let $V_{j,h} \in \{0,1\}$ denote the validity matrix, where $V_{j,h} = 1$ indicates that class $h$ is an admissible label for grid pixel $j$ under the mapping scheme in \ref{appendix:mappingscheme}. A prediction is considered correct at the sample level if it matches any element of the admissible label set, defined by the sample correctness indicator:
\begin{equation}
    \label{eq:sample_correctness}
    \tag{8}
    c_j = \mathbbm{1}\left(
    \sum_{h=1}^{k} P_{j,h}\, V_{j,h} > 0
    \right)
\end{equation}

Under this formulation, true positives, false positives, and false negatives for class $h$ are defined as:
\begin{align}
    \tag{9}
    TP_h &= \sum_{j=1}^{|M_i|} P_{j,h}\, V_{j,h} \label{eq:tp} \\
    \tag{10}
    FP_h &= \sum_{j=1}^{|M_i|} P_{j,h} \left(1 - V_{j,h}\right) \label{eq:fp} \\
    \tag{11}
    FN_h &= \sum_{j=1}^{|M_i|} \left(1 - c_j\right) V_{j,h} \label{eq:fn}
\end{align}

\noindent where $TP_h$ counts grid pixels correctly assigned to a valid class $h$, $FP_h$ counts grid pixels assigned to class $h$ when it is not admissible, and $FN_h$ counts grid pixels for which class $h$ is valid but the overall prediction is incorrect. The per-class F1-score and IoU are then computed as:
\begin{align}
    \tag{12}
    F1_h &= \frac{2\,TP_h}{2\,TP_h + FP_h + FN_h + \epsilon} \label{eq:f1} \\
    \tag{13}
    IoU_h &= \frac{TP_h}{TP_h + FP_h + FN_h + \epsilon} \label{eq:iou}
\end{align}

\noindent where $\epsilon$ is a small constant for numerical stability. Macro-averaged metrics are computed as the unweighted mean across all $k$ classes:
\begin{align}
    \tag{14}
    \overline{F1} = \frac{1}{k} \sum_{h=1}^{k} F1_h \\
    \tag{15}
    \overline{IoU} = \frac{1}{k} \sum_{h=1}^{k} IoU_h \label{eq:macro}
\end{align}

Frequency-weighted metrics aggregate per-class scores according to class frequency derived from $|C^{h}_{s_i}|$. We define the relative frequency of class $h$ in sector $s_i$ as:
\begin{equation}
    \tag{16}
    f^{h}_{s_i} = \frac{|C^{h}_{s_i}|}{\sum_{l=1}^{k} |C^{l}_{s_i}|}
\end{equation}

We used $w_h = f^{h}_{s_i}$ as the frequency weight. Frequency-weighted metrics are then defined as:
\begin{align}
    \tag{17}
    \overline{F1}^{w} &= \sum_{h=1}^{k} w_h\, F1_h \\
    \tag{18}
    \overline{IoU}^{w} &= \sum_{h=1}^{k} w_h\, IoU_h
    \label{eq:weighted}
\end{align}

To further assess learning behavior with respect to class frequency, minority classes were defined as classes with a sector-level relative frequency below 20\%, based on inspection of the observed class frequency distributions. Formally, a class $h$ is considered as a minority class in sector $s_i$, when $f^{h}_{s_i} < 0.20$.


\begin{table*}[tp]
    \centering
    \caption{Comparison of the total buildings, dwellings, and occupants among \textsc{DeepC4}, GEM, and METEOR. The percentage values inside the parentheses indicate the accuracy in percentage errors with respect to the census records.}
    \label{tab:byBuildingDwellingOccupant}
    \begin{tabular}{lrrrr}
    \hline
    \multicolumn{1}{c}{\textbf{Counts}} &
      \multicolumn{1}{c}{\textbf{Census}} &
      \multicolumn{1}{c}{\textbf{\textsc{DeepC4}}} &
      \multicolumn{1}{c}{\textbf{GEM}} &
      \multicolumn{1}{c}{\textbf{METEOR}} \\ \hline
    Building & -          & 3,200,582           & 3,258,527           & 3,368,644 \\
    Dwelling & 3,312,743  & \textbf{3,350,277 (1.13\%)}  & 3,379,883 (2.03\%)  & -         \\
    Occupant & 13,100,600 & \textbf{13,246,394 (1.11\%)} & 13,531,804 (3.29\%) & -         \\ \hline
    \end{tabular}
\end{table*}

\begin{table*}[htpb!]
    \centering
    \caption{Comparison of the total buildings, dwellings, and occupants in urban and rural settlements between \textsc{DeepC4} and GEM. The percentage values inside the parentheses indicate the accuracy in percentage errors with respect to the census records.}
    \label{tab:bySettlementType}
    \begin{tabular}{lrrrrrr}
    \cline{2-7}
             & \multicolumn{2}{c}{\textbf{Census}} & \multicolumn{2}{c}{\textbf{\textsc{DeepC4}}}                    & \multicolumn{2}{c}{\textbf{GEM}}      \\ \cline{2-7} 
     & \multicolumn{1}{c}{Rural} & \multicolumn{1}{c}{Urban} & \multicolumn{1}{c}{Rural} & \multicolumn{1}{c}{Urban} & \multicolumn{1}{c}{Rural} & \multicolumn{1}{c}{Urban} \\ \hline
    Building & -                & -                & 2,270,418                   & 929,866                  & 2,448,058          & 810,469          \\
    Dwelling & 2,348,456        & 964,287          & \textbf{2,368,775 (0.87\%)} & \textbf{981,502 (1.8\%)} & 2,536,213 (8\%)    & 843,670 (-13\%)  \\
    Occupant & 9,545,149        & 3,701,245        & \textbf{9,545,149 (0\%)}    & \textbf{3,701,245 (0\%)} & 10,075,690 (5.6\%) & 3,456,114 (-7\%) \\ \hline
    \end{tabular}
\end{table*}

\begin{table*}[htpb!]
    \centering
    \captionsetup{justification=centering}
    \caption{Comparison of the total buildings for each general macro-taxonomy class among \textsc{DeepC4}, GEM, and METEOR.}
    \label{tab:byWall}
    \begin{tabular}{l|rrr|ccc}
    \hline
    \textbf{Description} &
      \multicolumn{1}{c}{\textbf{\textsc{DeepC4} (A)}} &
      \multicolumn{1}{c}{\textbf{GEM (B)}} &
      \multicolumn{1}{c|}{\textbf{METEOR (C)}} &
      \textbf{A vs C} &
      \textbf{B vs C} &
      \textbf{A vs B} \\ \hline
    All                 & 3,200,582 & 3,258,527 & 3,368,644 & 5.1\%         & 3.3\%          & \textbf{1.8\%} \\
    Sundried blocks     & 1,594,922 & 1,340,170 & 1,813,852 & \textbf{13\%} & 30\%           & 17\%           \\
    Reinforced concrete & 20,017    & 16,920    & 32,158    & 47\%          & 62\%           & \textbf{17\%}  \\
    Informal            & 4,470     & 24,061    & 27,228    & 144\%         & \textbf{12\%}  & 137\%          \\
    Rubble stone        & 18,672    & 8,563     & 27,697    & \textbf{39\%} & 106\%          & 74\%           \\
    Concrete block      & 15,935    & 19,769    & 58,281    & 114\%         & 99\%           & \textbf{21\%}  \\
    Brick masonry       & 665,498   & 545,573   & 126,049   & 136\%         & 125\%          & \textbf{20\%}  \\
    Wood                & 24,393    & 22,736    & 28,832    & 17\%          & 24\%           & \textbf{7\%}   \\
    Wattle and daub     & 856,675   & 1,280,734 & 1,254,547 & 38\%          & \textbf{2.1\%} & 40\%           \\ \hline
    \end{tabular}
\end{table*}

\section{Results and Discussion}
\label{ResultsAndDiscussion}

\subsection{Comparing \textsc{DeepC4}, METEOR, and GEM}

Comprehensively comparing the outputs of \textsc{DeepC4}, METEOR, and GEM and disentangling the benefits of using \textsc{DeepC4} and improved fine-grained inputs (e.g., average household size for urban and rural areas) are challenging and non-trivial because of the closed-source information on the methodological implementation of past studies and their original input data. Thus, in the succeeding discussions, the basis of our comparative analyses and discussions is the resulting outputs as compared to the available 2022 census-based aggregated information in \autoref{tab:byBuildingDwellingOccupant}, \autoref{tab:bySettlementType}, and \autoref{tab:bySpatialDisaggregationDeepC4GEM}. In addition, because of the absence and inherent difficulty to maintain groundtruth information for other inferred outputs such as the distributions of general macro-taxonomy classes and height categories at the national level, we instead calculated pairwise percent differences to assess inter-model agreement among the outputs of \textsc{DeepC4}, METEOR, and GEM in \autoref{tab:byWall} and \autoref{tab:byHeight}.

Another key challenge is the temporal inconsistency among the outputs of \textsc{DeepC4}, METEOR, and GEM. Thus, we uniformly scaled up the outputs of METEOR and GEM using the multipliers, 1.37 and 1.05, to convert from 2012 and 2020, respectively, to 2022. These multipliers are based on the assumptions that the outputs of METEOR used the 2012 dwelling counts from the available census records during its time of analysis and that the outputs of GEM derived a 50-year linear trend for building counts. Although such a simple projection approach can further propagate uncertainties, the corresponding converted outputs of METEOR and GEM were found to be comparable, with the smallest possible percentage differences, confirming that these assumptions can be used reliably. We also compare the outputs of \textsc{DeepC4} with those of METEOR and GEM in terms of their respective numerical estimates by urban morphology indicators and the quality of their spatial disaggregation, and discuss their respective accuracies, for cases wherein census records are available. 

\subsubsection{By building, dwelling, and occupant counts}

At the national level, our findings in \autoref{tab:byBuildingDwellingOccupant} show that the numbers of dwellings and occupants of the outputs of \textsc{DeepC4} have higher accuracy with percentage errors of 1.13\% and 1.11\% than those of GEM with percentage errors of 2.03\% and 3.29\%, respectively. For the number of buildings, the outputs of \textsc{DeepC4} also reveal increasing differences with those of GEM ($\sim$58k more) and METEOR ($\sim$168k more), which also follows that the corresponding numbers of dwellings and occupants of METEOR will have the lowest accuracy than those of \textsc{DeepC4} and GEM. While the linear scaling between two temporal periods of measurement could provide a possible projection of the trend in the number of buildings, our results imply that the Rwandan urban development through the years may have followed a more complex dynamics, indicating that linear scaling may lead to an overestimation of the number of buildings, dwellings, and occupants.

When classified into settlement types, in \autoref{tab:bySettlementType}, the numbers of rural and urban dwellings of the outputs of \textsc{DeepC4} result in percentage errors of 0.87\% and 1.8\%, respectively, whereas those of GEM overestimate the rural dwelling by 8\% and underestimate the urban dwelling by 13\%. Similarly, both rural and urban occupants considered by the \textsc{DeepC4} perfectly match the official census records, whereas those of GEM respectively deviated by 5.6\% and -7\%. Considering the number of buildings, the outputs of \textsc{DeepC4} estimated $\sim$178k fewer rural buildings and $\sim$119k more urban buildings.

This signifies the value of a more fine-grained census statistics at the level of sectors, effectively reducing the resulting uncertainty across resolutions, compared to GEM, which used a single average household size for the urban and rural areas of the entire Rwanda and five rows of data on population counts at the provincial level \citep{paul2022development}. In addition, these findings also underscore the effect of the selection of the average household size at different levels of aggregation on the reliability and uncertainty of the resulting large-scale exposure and risk outputs, which have been shown to be important in characterizing the urban-rural divide and its relative regional degrees of physical vulnerability for any subsequent risk calculation.

\subsubsection{By urban morphology indicators}

Due to their respective inherent modelling differences, it is difficult to perform a perfect one-to-one similarity on the inferred distribution of urban morphology among \textsc{DeepC4}, METEOR, and GEM. Thus, instead, we provide a weak comparison on the general groups of macro-taxonomy classes in \autoref{tab:byWall} using the assumed relationships in Appendix \ref{appendix:mappingscheme}. In \autoref{tab:byHeight}, we also present a comparison of the distribution of the classes of building height between \textsc{DeepC4} and GEM to demonstrate the incorporation of an expert belief system in both approaches.

\begin{table}[tp]
    \centering
    \caption{Comparison of the total buildings for each height category between \textsc{DeepC4} and GEM.}
    \label{tab:byHeight}
    \begin{tabular}{lrrc}
    \hline
    \textbf{Class} & \multicolumn{1}{c}{\textbf{\textsc{DeepC4}}} & \multicolumn{1}{c}{\textbf{GEM}} & \textbf{\% Difference} \\ \hline
    All      & 3,200,284 & 3,258,527 & 1.8\% \\
    H:1      & 1,776,985 & 1,927,411 & 6.9\% \\
    H:2      & 1,397,280 & 1,312,061 & 4.7\% \\
    H:3      & 10,380    & 6,825     & 17\%  \\
    HBET:3-6 & 15,276    & 12,048    & 18\%  \\
    HBET:4-7 & 291       & 144       & 23\%  \\
    HBET:8+  & 73        & 38        & 13\%  \\ \hline
    \end{tabular}
\end{table}

In \autoref{tab:byWall}, while the total buildings across all general groups of macro-taxonomy classes observe minimal percentage differences of 1.8\% to 5.1\%, \textsc{DeepC4} and GEM, on one hand, share a closer agreement for reinforced concrete, concrete block, brick masonry, and wood. On the other hand, \textsc{DeepC4} and METEOR have more precise estimates for sundried blocks and rubble stones, whereas, only under the categories of informal and wattle-and-daub, \textsc{DeepC4} is less favorable when compared to GEM and METEOR. 

Despite the absence of building-level information to validate this distribution of macro-taxonomy classes, our results still highlight the observed sensitivity on the possible composition of macro-taxonomy classes of the building stock at a large scale using various approaches. Nevertheless, \autoref{tab:byWall} shows that outputs of \textsc{DeepC4} and GEM have a more similar underlying expert belief system, except for the 'informal' category, which can be attributed to the overlapping with other macro-taxonomy classes and the use of sector-level rural and urban household sizes, significantly affecting the allocation of the number of buildings under the said category in \textsc{DeepC4}.  

For the inferred distribution of building heights between \textsc{DeepC4} and GEM, \autoref{tab:byHeight} shows that the more discrete height categories (i.e., H:1, H:2, and HBET:8+) achieve the lowest percentage differences up to 13\%, while the categories of building heights with overlapping ranges (i.e., H:3, HBET:3-6, and HBET:4-7) have higher percentage differences up to 23\%. When combined, the group of H:3, HBET:3-6, and HBET:4-7 corresponds to a total percentage difference of 31\%, with \textsc{DeepC4} estimating 6,930 more buildings. In addition, \textsc{DeepC4} has higher estimates for taller buildings (i.e., H:2, H:3, HBET:3-6, and HBET:4-7) while GEM has only higher estimates for H:1.

While \textsc{DeepC4} incorporated the weak supervision from the available building-level height information from \citet{bachofer2019building}, these results also reflect the uncertainty in associating lower and upper bounds of building heights with a specific number of floor levels, as shown in \autoref{appendixtab:buildingheightmapping} in \ref{appendix:mappingscheme}. For example, H:1 may likely correspond to a building height of 3 meters, but, due to the uniqueness of the different wall types and macro-taxonomy classes, it can possibly be lower for the case of many informal settlements with weak structural integrity and higher for the case of more improved structural designs of buildings, such as reinforced concrete. 

Our results, however, provide an alternative distribution of building height classes that is both based on the expert belief system also used in \citet{paul2022development} and an extended benchmarking from the use of weak supervision from the available groundtruth data from \citet{bachofer2019building}. Overall, we note that integrating expert belief systems (e.g., local understanding of the number of floor levels) and observational data (e.g., mapped building heights in meters) is a key challenge in characterizing the regional urban morphology for disaster risk assessment. 

\subsubsection{By spatial disaggregation quality}

\begin{figure*}[t]
    \centering
    \makebox[\textwidth][c]{\includegraphics[width=7.2in]{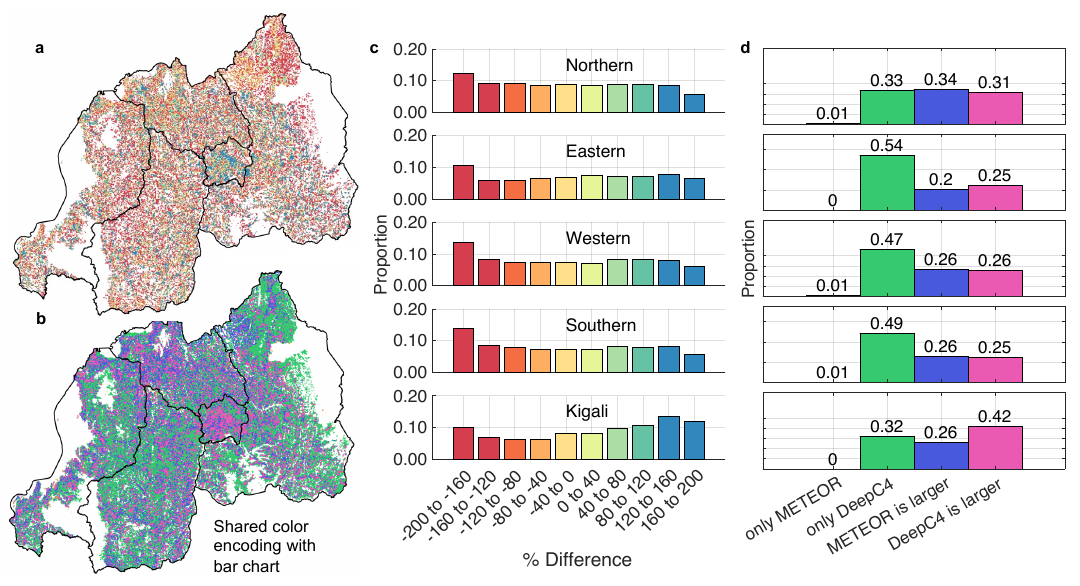}}
    \caption{Comparison of the spatial disaggregation between \textsc{DeepC4} and METEOR at 500-meter pixel. Calculated percentage differences \textbf{(a)} visualized on a map and \textbf{(c)} expressed as a histogram chart showing the proportion of pixels per province. Positive percentage differences indicate that outputs of \textsc{DeepC4} are larger than those of METEOR, and negative otherwise. Categorical difference as \textbf{(b)} a map and \textbf{(d)} a histogram chart showing the proportion of pixels for each descriptive category. The color encoding of the map and bar chart is shared between \textbf{(a)} and \textbf{(c)}, and between \textbf{(b)} and \textbf{(d)}.}
    \label{fig:bySpatialDisaggregationDeepC4METEOR}
\end{figure*}

Because of the differences in spatial resolutions of aggregation, we present two comparative analyses: \autoref{fig:bySpatialDisaggregationDeepC4METEOR} shows a map and histogram of percentage differences between \textsc{DeepC4} and METEOR at the 500-meter pixel; and \autoref{tab:bySpatialDisaggregationDeepC4GEM} presents percentage differences on the number of dwellings considered by \textsc{DeepC4} and GEM and their corresponding accuracy in percentage errors with respect to the official census records at the province level. 

\begin{table*}[t]
    \centering
    \caption{Comparison of the spatial disaggregation between \textsc{DeepC4} and GEM at the province level. For each province, this shows the number of urban, rural, and total dwellings. The percentage values inside the parentheses indicate the accuracy with respect to the census records.}
    \label{tab:bySpatialDisaggregationDeepC4GEM}
    \begin{tabular}{llrrrc}
    \hline
    \textbf{Dwelling} &
      \textbf{Province} &
      \multicolumn{1}{c}{\textbf{Census}} &
      \multicolumn{1}{c}{\textbf{\textsc{DeepC4}}} &
      \multicolumn{1}{c}{\textbf{GEM}} &
      \textbf{\% Difference} \\ \hline
    \multirow{5}{*}{Total} & North  & 506,064 & \textbf{509,744 (0.7\%)} & 483,807 (-4.4\%) & 5.22\%  \\
                           & East   & 886,132 & \textbf{896,945 (1.2\%)} & 950,320 (7.2\%)  & 5.78\%  \\
                           & West   & 671,506 & \textbf{677,394 (0.9\%)} & 733,609 (9.2\%)  & 7.97\%  \\
                           & South  & 760,173 & \textbf{770,372 (1.3\%)} & 790,089 (3.9\%)  & 2.53\%  \\
                           & Kigali & 488,868 & \textbf{495,822 (1.4\%)} & 422,058 (-14\%)  & 16.07\% \\ \hline
    \multirow{5}{*}{Rural} & North  & 417,670 & \textbf{419,807 (0.5\%)} & 403,373 (-3.4\%) & 3.99\%  \\
                           & East   & 700,049 & \textbf{705,767 (0.8\%)} & 792,306 (13\%)   & 11.55\% \\
                           & West   & 522,847 & \textbf{527,834 (1.0\%)} & 611,632 (17\%)   & 14.71\% \\
                           & South  & 651,454 & \textbf{654,395 (0.5\%)} & 658,727 (1.1\%)  & 0.66\%  \\
                           & Kigali & 56,436  & \textbf{60,973 (8.0\%)}  & 70,176 (24\%)    & 14.03\% \\ \hline
    \multirow{5}{*}{Urban} & North  & 88,394  & \textbf{89,937 (1.7\%)}  & 80,434 (-9.0\%)  & 11.16\% \\
                           & East   & 186,083 & \textbf{191,178 (2.7\%)} & 158,014 (-15\%)  & 18.99\% \\
                           & West   & 148,659 & \textbf{149,560 (0.6\%)} & 121,977 (-18\%)  & 20.32\% \\
                           & South  & 108,719 & \textbf{115,978 (6.7\%)} & 131,362 (21\%)   & 12.44\% \\
                           & Kigali & 432,432 & \textbf{434,849 (0.6\%)} & 351,882 (-19\%)  & 21.09\% \\ \hline
    \end{tabular}
\end{table*}

Although METEOR estimated $\sim$168k more buildings, \autoref{fig:bySpatialDisaggregationDeepC4METEOR}b and ~\ref{fig:bySpatialDisaggregationDeepC4METEOR}d show that outputs of \textsc{DeepC4} occupy more 500-meter grid pixels by 32\% (Kigali City) to 49\% (Southern Province), which is indicative of the urban development captured by the EO data for the year 2022. While METEOR estimates a slightly higher number of buildings in 34\% of all grid pixels in the Northern Province, the general trend across all provinces in \textsc{DeepC4} still shows that 25\% of all grid pixels in Eastern and Southern Provinces and 42\% for Kigali City have more building counts than METEOR. 

It is important to note that \autoref{fig:bySpatialDisaggregationDeepC4METEOR}b and ~\ref{fig:bySpatialDisaggregationDeepC4METEOR}d illustrate only the differences in the space or grid pixels with which either \textsc{DeepC4} or METEOR has disaggregated values, regardless of the magnitude of their differences. However, these results emphasize that the quality of large-scale exposure datasets also concerns not only the numerical accuracy of the disaggregated building counts but also the spatial accuracy as to where these buildings are located on the map, signifying the critical influence of urban dynamics and temporal consistency of EO data on their degrees of exposure to natural hazards.

Furthermore, in terms of the magnitude of differences between \textsc{DeepC4} and METEOR, \autoref{fig:bySpatialDisaggregationDeepC4METEOR}c shows that \textsc{DeepC4} and METEOR result in uniformly distributed percentage differences. However, towards the left end or colored red in \autoref{fig:bySpatialDisaggregationDeepC4METEOR}c, it also indicates that the outputs of METEOR are significantly higher than those of \textsc{DeepC4} for all surrounding and larger provinces except Kigali City, despite METEOR occupying fewer grid pixels than \textsc{DeepC4} in the previous discussion of \autoref{fig:bySpatialDisaggregationDeepC4METEOR}d. \autoref{fig:bySpatialDisaggregationDeepC4METEOR}a also confirms that the province of Kigali City has more grid pixels that are colored blue or higher building counts from \textsc{DeepC4}, which is equivalent to the observed distribution of positive percentage differences in \autoref{fig:bySpatialDisaggregationDeepC4METEOR}c. 

These results imply that both the size or areal extent and the relative location (i.e., whether the region is located in the center or perimeter) affect the quality of spatial disaggregation. Particularly, the large areal extents of Northern, Eastern, Western, and Southern Provinces provide more possible and uncertain locations for disaggregation, while the province of Kigali City with the lowest areal extent at the center has limited and constrained locations, resulting in the lowest categorical difference at 32\% where an estimate of \textsc{DeepC4} exists in \autoref{fig:bySpatialDisaggregationDeepC4METEOR}d.

In \autoref{tab:bySpatialDisaggregationDeepC4GEM}, even at the province level, \textsc{DeepC4} appears to have a set of more accurate numbers of dwellings for both urban and rural areas than those of GEM. For urban dwellings, GEM underestimates by 9\% to 19\% for all provinces except the South Province, which observes a large overestimation by 21\%. For rural dwellings, GEM significantly overestimates Kigali City by 24\%, followed by the East and West Provinces by 13\% and 17\%, respectively. Nevertheless, GEM achieves the most accurate numbers for rural dwellings for the North and South provinces, which are also the most precise estimates with the least percentage differences at 3.99\% and 0.66\%, respectively, when compared with the outputs of \textsc{DeepC4}.

Our results show that constraining the spatial disaggregation by incorporating census statistics and encoding expert belief systems in a form of probabilistic relationships among various urban morphology indicators has effectively established a consensus while learning efficient reduced-dimension representations that are able to cluster under weak supervision of available building-level groundtruth information and infer the distribution of the characteristics of the urban layout.

\subsection{Assessing Overall Predictive Performance}

\begin{figure}[b!]
    \centering
    \includegraphics[width=85mm]{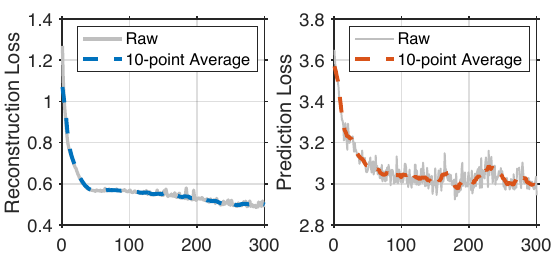}
    \caption{Training loss evolution over iterations.}
    \label{fig:ModelPerformance}
\end{figure}

To understand the reliability of the learning capability and predictive performance of our trained \textsc{DeepC4} model, we present the learning curves in \autoref{fig:ModelPerformance} with an example set of predictions and $p_{\mathrm{valid}}$ maps in \autoref{fig:pValidMap} on the use of weak supervision of partially available building-level groundtruth of \citet{bachofer2019building}. \autoref{fig:ModelPerformance} shows that our trained \textsc{DeepC4} model has converged to a set of stable reduced-dimension latent representations, resulting in a mean-squared error of 0.50 (a 53\% reduction from the initially calculated reconstruction loss) and a Euclidean distance-based prediction loss of 2.99 (a measurable 16\% reduction from the initially calculated prediction loss). 

In addition, the observed fluctuation trends of prediction loss in \autoref{fig:ModelPerformance} \textit{(right)}, which are more pronounced than those of the reconstruction loss in \autoref{fig:ModelPerformance} \textit{(left)}, can be attributed to the varying sizes and non-uniform clustering behavior of sectors with different proportions of clusters. In contrast, the smoother curve of reconstruction loss is inherent to the spatially continuous structure of EO satellite imagery signals, which vary gradually across neighboring pixels. 

In the succeeding sections, we provide a comprehensive evaluation of the predictive performance of \textsc{DeepC4}. \autoref{sec:metrics} introduces the validity-based classification metrics, reporting standard macro-averaged and frequency-weighted F1 and IoU alongside $p_{\mathrm{valid}}$ to enable benchmarking. \autoref{sec:ablation} presents an ablation study on the contribution of each input modality, empirically validating the necessity of multimodal fusion from each task. \autoref{sec:minority} examines minority class learning trajectories to assess whether the model jointly learns discriminative representations beyond spatial priors. \autoref{sec:lambda} reports a sensitivity analysis between reconstruction and prediction losses to verify the robustness of the unweighted formulation. Finally, \autoref{sec:spatial} presents an example set of prediction maps and their spatial distributions of predictive performance through $p_{\mathrm{valid}}$, complementing the aggregate metrics reported in the preceding sections.
 
\subsubsection{Validity-based classification metrics}
\label{sec:metrics}

\begin{table*}[b]
    \centering
    \caption{Classification metrics computed under the validity-based formulation at the optimal epoch of 300. Macro-averaged metrics treat all classes equally, while frequency-weighted metrics aggregate per-class scores according to class frequency.}
    \label{tab:classificationMetrics}
    \setlength{\tabcolsep}{8pt}
    \renewcommand{\arraystretch}{1.2}
    \begin{tabular}{lcccccc}
    \hline
    & \multicolumn{2}{c}{\textbf{Macro}} & \multicolumn{2}{c}{\textbf{Frequency-weighted}} & \multicolumn{2}{c}{\textbf{Validity-based}} \\
    \cline{2-3} \cline{4-5} \cline{6-7}
    \textbf{Task} & \textbf{F1} & \textbf{mIoU} & \textbf{F1} & \textbf{mIoU} & \textbf{$p_{\mathrm{valid}}$} & \textbf{$p_{\mathrm{valid}}^{w}$} \\
    \hline
    Roof   & 0.629 & 0.575 & 0.991 & 0.985 & 0.991 & 0.996 \\
    Wall   & 0.784 & 0.711 & 0.955 & 0.922 & 0.969 & 0.578 \\
    Height & 0.451 & 0.417 & 0.956 & 0.927 & 0.965 & 0.114 \\
    \hline
    \end{tabular}
\end{table*}

\autoref{tab:classificationMetrics} reports standard classification metrics computed under the validity-based formulation at the optimal epoch. In this setting, a prediction is considered correct if it matches any element of the admissible label set, from which per-class F1 and IoU are derived and aggregated as both macro-averaged and frequency-weighted scores. The full multimodal configuration achieves a macro-F1 of 0.629 and macro-mIoU of 0.575 for roof, 0.784 and 0.711 for wall, and 0.451 and 0.417 for height. The combination of macro and frequency-weighted metrics provides a transparent evaluation under both class-balanced and real-world frequency distributions, facilitating comparison with future work under similar weak supervision settings.

The significant frequency-weighted performance across all tasks, particularly for roof (F1: 0.991, mIoU: 0.985), reflects the influence of class imbalance wherein a small number of classes dominate the majority of grid pixels as previously indicated in \autoref{tab:byHeight}, in addition to the overlapping mapping scheme between the building-level groundtruth information and the desired indicators of urban morphology. The discrepancy between macro and frequency-weighted metrics for height, where macro-F1 is 0.451 yet frequency-weighted F1 reaches 0.956, further confirms that dominant height classes are predicted well while minority classes remain challenging. This is consistent with the low \textbf{$p_{\mathrm{valid}}^{w}$} of 0.114 for height, suggesting that the number of valid label combinations for minority height classes may be limited and overlapping despite our efforts in linking different classifications between \citet{bachofer2019building} and the classes in \autoref{table:classes}.

The validity-based \textbf{$p_{\mathrm{valid}}$} reached 0.991, 0.969, and 0.965 for roof, wall, and height, respectively, with the corresponding frequency-weighted \textbf{$p_{\mathrm{valid}}$} reported in \autoref{tab:classificationMetrics}. The markedly lower \textbf{$p_{\mathrm{valid}}$} for wall (0.578) and height (0.114) relative to their absolute \textbf{$p_{\mathrm{valid}}$} values reinforces that dominant classes drive aggregate performance, which is a common challenge in urban morphology disaggregation tasks with overly dominating groups of classes. This also implies that the optimality based on the imbalanced number of target classes and different standards for target classes is an important consideration in spatial disaggregation. Nevertheless, even though the partially available building-level groundtruth provides limited or weak supervision, the results of our 5-fold cross-validation implementation show a drop across all metrics by only 2\%.

\subsubsection{Ablation study on multimodal input contribution}
\label{sec:ablation}

\begin{figure*}[t]
    \centering
    \makebox[\textwidth][c]{\includegraphics[width=7.4in]{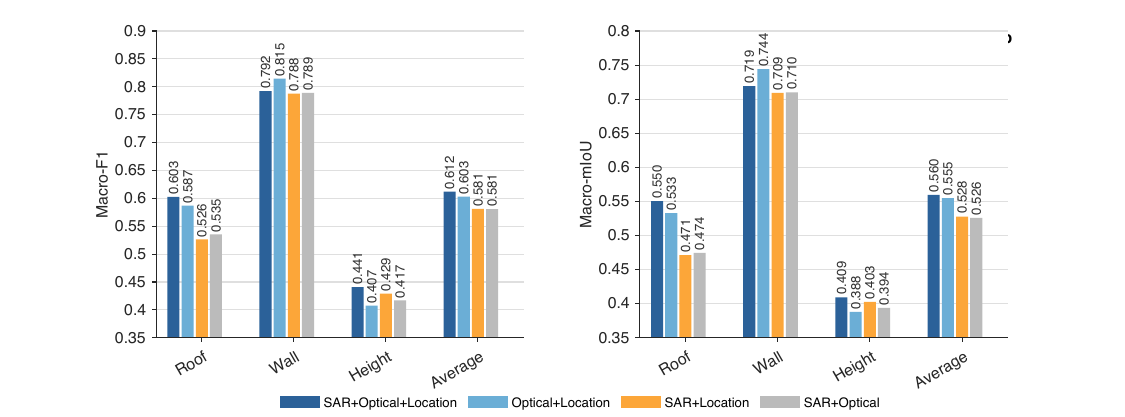}}
    \caption{Ablation study results comparing the full multimodal configuration (SAR + Optical + Location) against three ablated variants across roof, wall, height, and their average, assessed at epoch 100. Macro-F1 \textit{(left)} and macro-mIoU \textit{(right)} scores under the validity-based formulation. Higher values indicate better performance.}
    \label{fig:AblationStudy}
\end{figure*}

To empirically validate the necessity of each input modality, we conducted an ablation study by independently removing SAR (Sentinel-1 SAR GRD) and optical imagery (Sentinel-2 Harmonized MSI) while keeping the remaining architecture and training configuration unchanged, all assessed at epoch 100. \autoref{fig:AblationStudy} presents macro-F1 and macro-mIoU for four configurations: the full multimodal model (SAR + Optical + Location), and three ablated variants. The complete set of metrics, including frequency-weighted F1, mIoU, and validity-based scores, is reported in the supplementary file.

Across all tasks, the full multimodal configuration achieved the highest average macro-F1 (0.612) and macro-mIoU (0.560), with the modality contributions being task-specific and physically interpretable. Optical imagery is the dominant signal for roof prediction, as the surface spectral properties are directly observable from top-down satellite imagery \citep{braun2019}. For the roof prediction task, its removal reduced macro-F1 and macro-mIoU by -0.077 and -0.079, respectively, a relative drop of -12.8\% and -14.4\%, compared to a marginal drop by -2.65\% and -3.09\% when SAR is removed. Conversely, SAR provides the dominant contribution for height prediction, as the backscatter intensity contains structural and volumetric building properties \citep{koppel2017}. For the height prediction task, its removal reduced macro-F1 and macro-mIoU by -0.034 and -0.021, respectively, approximately threefold the performance drop observed when removing optical imagery (-0.012 and -0.006). For the wall prediction task, optical imagery contributes positively while SAR introduces a marginal trade-off, with the Optical+Location variant achieving a marginally higher macro-F1 (0.815 vs. 0.792) and macro-mIoU (0.744 vs. 0.719), which is consistent with the physical limitation of inferring side-view wall material properties from top-down satellite observations. Overall, no single ablated configuration achieves the best performance across all tasks simultaneously. The full multimodal model achieves the highest average macro-F1 and macro-mIoU by leveraging complementary and task-specific contributions from each modality within the joint multitask framework.

Frequency-weighted metrics and \textbf{$p_{\mathrm{valid}}$} were more stable across configurations, reflecting the robustness of dominant class predictions to modality removal. The full multimodal configuration nonetheless retained the highest average frequency-weighted F1 (0.966), frequency-weighted mIoU (0.943), and absolute \textbf{$p_{\mathrm{valid}}$}  (0.974). The \textbf{$p_{\mathrm{valid}}^{w}$} exhibited greater variability, particularly for height, consistent with the sensitivity of minority classes. Overall, these results support the necessity of multimodal fusion, demonstrating that SAR and optical imagery provide complementary and physically interpretable contributions across tasks.

\subsubsection{Multimodal contribution to minority class learning}
\label{sec:minority}

\begin{figure*}[t]
    \centering
    \makebox[\textwidth][c]{\includegraphics[width=7.4in]{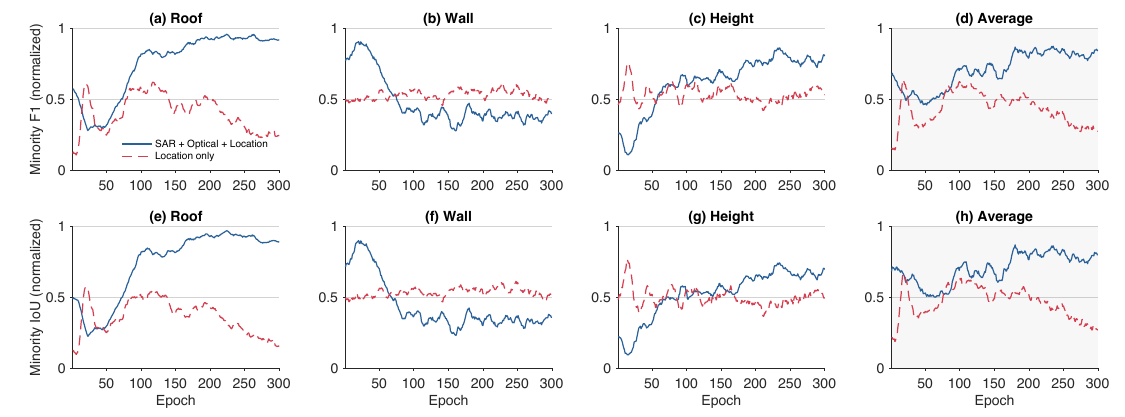}}
    \caption{Normalized mean minority class F1 \textit{(top row)} and IoU \textit{(bottom row)} over 300 training epochs for the full multimodal model (SAR + Optical + Location) \textit{(solid blue line)} and the `Location only' \textit{(dashed red line)}, across roof, wall, height, and their average \textit{(from left to right)}.}
    \label{fig:minority}
\end{figure*}

We further examined how improvements from multimodal fusion are distributed between minority classes and dominant class predictions. For this analysis, we considered the `Location only' baseline as a prior proxy under the same cluster-level constraints, without any satellite imagery to learn from and relying solely on two location encoding features. \autoref{fig:minority} presents the normalized mean minority F1 and minority IoU over 300 training epochs for the full multimodal model and the `Location only' baseline, across roof, wall, height, and their average. Each panel is normalized to the combined range of both configurations to enable direct comparison across tasks with different absolute performance levels. The 'Location only' baseline exhibits near-horizontal and declining minority trajectories across all tasks, confirming its behavior as a prior baseline under the same cluster-level constraints. In contrast, the full multimodal model shows distinct non-flat learning dynamics throughout training by exploiting the rich information of EO data. For the roof prediction task, minority F1 and IoU increase consistently and diverge clearly upward from the `Location only' baseline after approximately epoch 50, confirming that optical imagery provides discriminative signals for rare roof classes beyond what spatial priors alone can capture. For the height prediction task, the full model similarly exhibits an upward trend and remains above the `Location only' baseline throughout training, reflecting the complementary contribution of SAR signals to minority height class discrimination.

For the wall prediction task, the full model exhibits a non-monotonic trajectory with an initial adjustment phase followed by stabilization, while the `Location only' baseline remains flat throughout. This observed behavior is consistent with the joint reorganization of shared latent representations across roof, wall, and height tasks during training, rather than prior-dominated learning. The flat `Location only' baseline confirms that spatial priors alone reach a performance ceiling for minority wall classes and cannot improve further. In contrast, the full model continues to adjust its latent representations, reflecting the physical limitation of inferring side-view wall material properties from top-down satellite observations. As also established in \autoref{sec:ablation}, no single ablated configuration achieves the best performance across all tasks simultaneously. 

Within the joint multitask framework, the average panel (see \autoref{fig:minority}d and ~\ref{fig:minority}h) confirms that the full multimodal model maintains consistently higher normalized mean minority F1 and IoU across all tasks combined throughout training, while the `Location only' baseline plateaus and declines. This observed difference is consistent with the effect of cluster-level constraints, which impose upper bounds on feasible class allocations across the labels of \citet{bachofer2019building}, rather than encoding probability distributions. Overall, these results collectively demonstrate that the model effectively learns minority class representations through multimodal integration, rather than fitting the prior probability distribution of the dominant classes.

\subsubsection{Loss weighting sensitivity analysis}
\label{sec:lambda}

To evaluate the robustness of the unweighted loss formulation adopted in this study ($\lambda = 1.0$), we performed a sensitivity analysis using $\lambda \in \{0.25, 0.50, 1.0, 2.0, 4.0\}$ under identical training settings. \autoref{tab:lambda} reports the macro-F1 and macro-mIoU for each configuration across roof, wall, height, and their average, while the complete set of metrics is provided in the supplementary material.

The observed performance remains largely stable across all tested values of $\lambda$, indicating low sensitivity in both macro-F1 and macro-mIoU. The unweighted formulation ($\lambda\!=\!1.0$) achieves the highest average macro-F1 (0.612) and macro-mIoU (0.560), together with the highest average $p_{\mathrm{valid}}$ (0.974) and frequency-weighted F1 (0.966). Although $(\lambda\!=\!2.0)$ produces the highest wall macro-F1 (0.803) and $(\lambda\!=\!4.0)$ produces the highest roof macro-F1 (0.625), the gains relative to $(\lambda\!=\!1.0)$ are small and do not translate into improved average performance. These results indicate that the framework is not highly sensitive to the relative weighting of clustering and reconstruction losses, supporting the use of the unweighted formulation as a simple and effective default configuration.

\begin{table*}[t]
    \centering
    \caption{Macro-F1 and macro-mIoU across roof, wall, height, and 
    their average for different $\lambda$ values, assessed at epoch 100.}
    \label{tab:lambda}
    \setlength{\tabcolsep}{6pt}
    \renewcommand{\arraystretch}{1.2}
    \begin{tabular}{ccccccccc}
    \hline
    & \multicolumn{2}{c}{\textbf{Roof}} 
    & \multicolumn{2}{c}{\textbf{Wall}} 
    & \multicolumn{2}{c}{\textbf{Height}} 
    & \multicolumn{2}{c}{\textbf{Average}} \\
    \cline{2-3} \cline{4-5} \cline{6-7} \cline{8-9}
    \textbf{$\lambda$} 
    & \textbf{F1} & \textbf{mIoU} 
    & \textbf{F1} & \textbf{mIoU} 
    & \textbf{F1} & \textbf{mIoU} 
    & \textbf{F1} & \textbf{mIoU} \\
    \hline
    0.25 & 0.620 & 0.574 & 0.763 & 0.680 & 0.440 & 0.409 & 0.608 & 0.554 \\
    0.50 & 0.622 & 0.569 & 0.731 & 0.643 & 0.418 & 0.395 & 0.590 & 0.536 \\
    \textbf{1.00} & 0.603 & 0.551 & 0.792 & 0.719 & 0.441 & 0.409 & \textbf{0.612} & \textbf{0.560} \\
    2.00 & 0.593 & 0.540 & 0.803 & 0.728 & 0.416 & 0.392 & 0.604 & 0.553 \\
    4.00 & 0.625 & 0.575 & 0.738 & 0.648 & 0.427 & 0.399 & 0.597 & 0.541 \\
    \hline
    \end{tabular}
\end{table*}

\begin{figure*}[t!]
    \centering
    \makebox[\textwidth][c]{\includegraphics[width=7.4in]{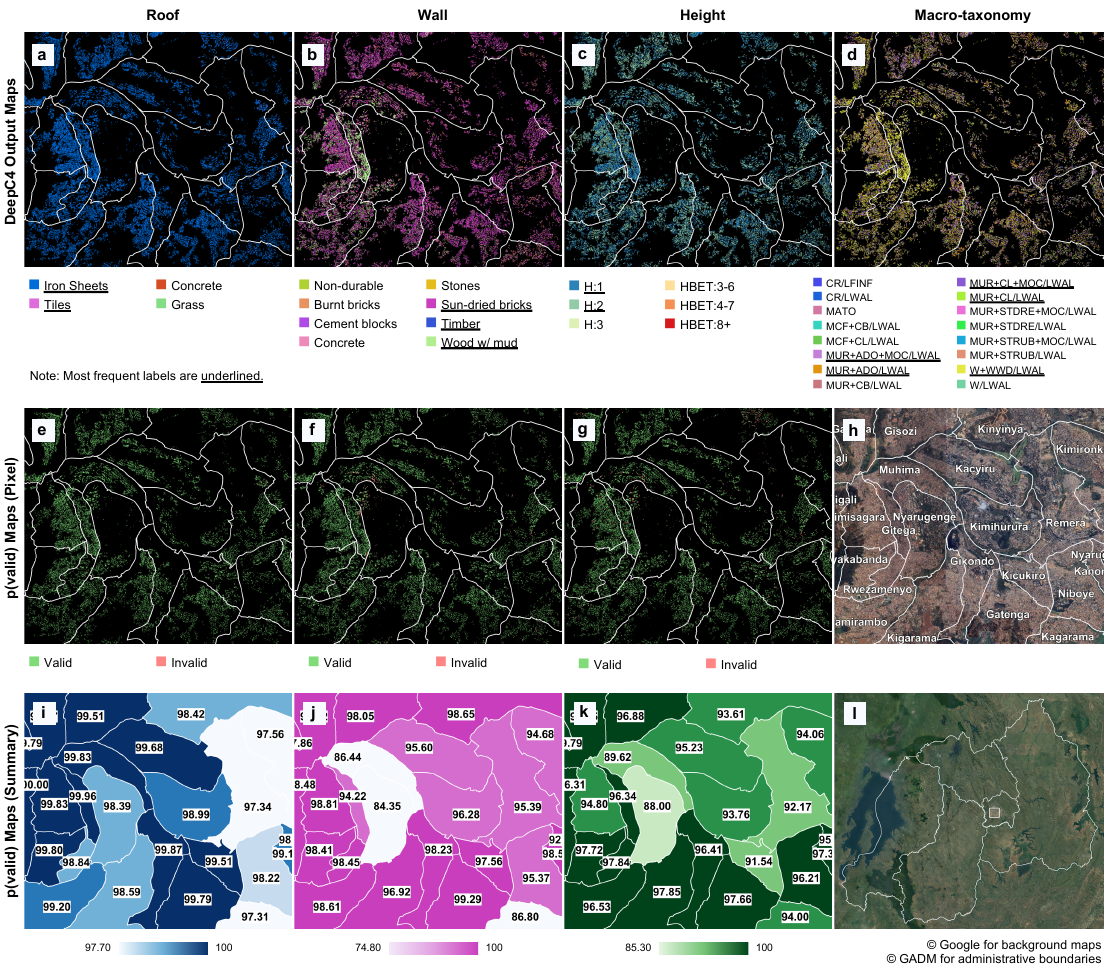}}
    \caption{Usefulness of partially available building-level groundtruth. \textbf{(a-d)} a set of prediction maps as a result of \textsc{DeepC4}; \textbf{(e-g)} corresponding $p_{valid}$ maps; \textbf{(h)} reference background map with sector labels; \textbf{(i-k)} aggregated sector-level values of $p_{valid}$ maps; and \textbf{(l)} reference location.}
    \label{fig:pValidMap}
\end{figure*}

\subsubsection{Spatial distribution of predictive performance}
\label{sec:spatial}
To visualize these metrics on the map, \autoref{fig:pValidMap}a to  ~\ref{fig:pValidMap}d show a set of prediction maps for the distribution of roof, wall, height, and macro-taxonomy classes over a few sectors in Kigali City, as labeled in Figure \ref{fig:pValidMap}h and referenced in Figure \ref{fig:pValidMap}l. Using the available building-level groundtruth information from which we trained our model using their calculated losses, \autoref{fig:pValidMap}e to ~\ref{fig:pValidMap}g shows that most grid pixels are accurately predicted (i.e., valid predictions in colored green) for maps of roof, wall, and height. Presenting these pixel maps as sector-level summaries in \autoref{fig:pValidMap}i to  ~\ref{fig:pValidMap}k, our results also confirm that our trained \textsc{DeepC4} model achieved a desirably improved performance with the clustering task for roof and height classes being notably more precise compared to that for wall.

\begin{figure*}[t]
    \centering
    \makebox[\textwidth][c]{\includegraphics[width=7.4in]{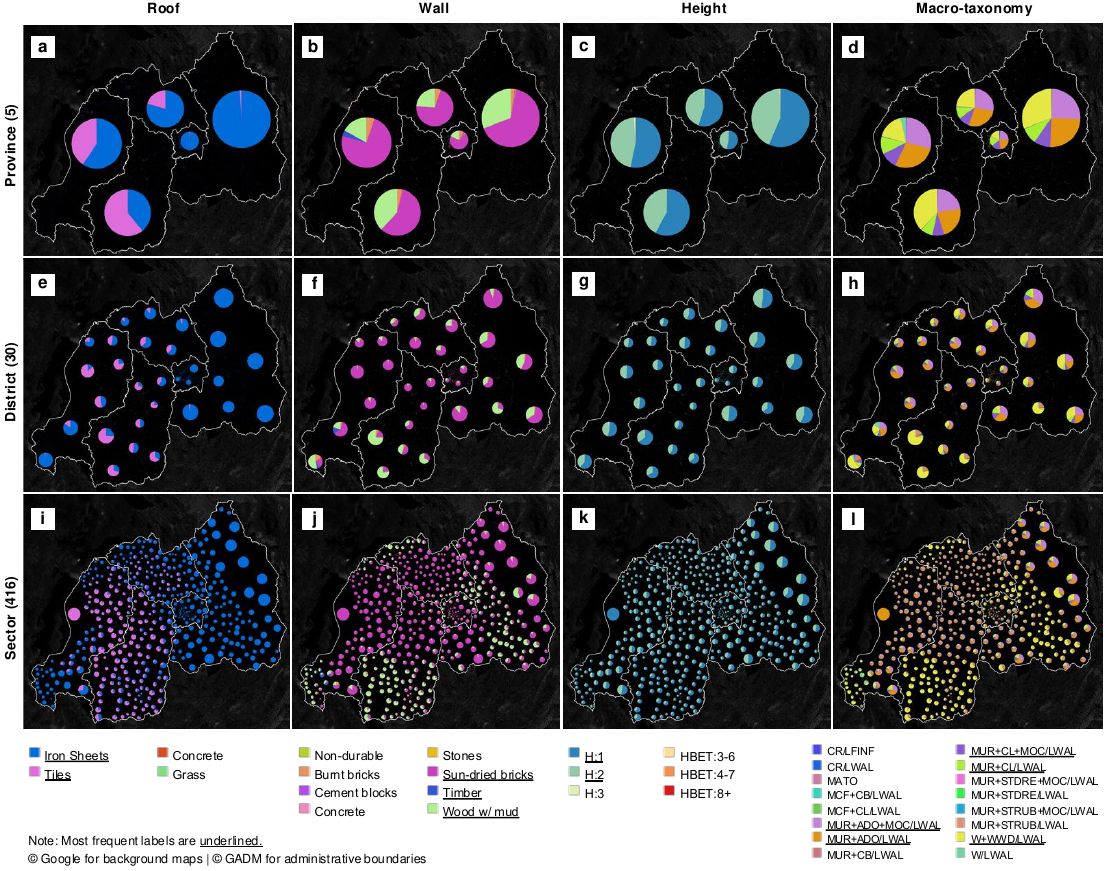}}
    \caption{Summary diagrams of the number of buildings for each urban morphology indicator (roof, wall, height, and macro-taxonomy classification, respectively from left to right) at the: \textbf{(a-d)} province, \textbf{(e-h)} district, and \textbf{(i-l)} sector level}
    \label{fig:summaryMaps}
\end{figure*}

\subsection{Mapping Rwandan Urban Morphology}

Figure \ref{fig:summaryMaps} summarizes the resulting prediction pixel maps into pie charts aggregated at the province, district, and sector levels. While the charts for roof and wall merely reflect the proportion dictated by the official census records because the clustering occurs at the sector level, \textsc{DeepC4} gives additional significant insights into the consistency of the quality of constraints for roof and wall and the distribution of height and macro-taxonomy class and extends the approach of the METEOR and GEM by performing a fine-grained implementation using the information derived from various EO signals and weakly supervised by the available building-level groundtruth data.

\autoref{fig:summaryMaps}a shows that roofs made of tiles are more prominent in western Rwanda, whereas iron sheet is the most commonly used roof material in eastern Rwanda. At the district level, we see in Figure \ref{fig:summaryMaps}e that tiles are more common in the Southern Province and on the northern portion of the Western Province. At the sector level in \autoref{fig:summaryMaps}i, the use of iron sheets evidently dominates the Eastern Province and Kigali City, including the southern part of the Western Province and the perimeter to the north of the Northern Province.

\autoref{fig:summaryMaps}b shows that the majority of wall material is sun-dried bricks, followed by wood with mud. At the district level in \autoref{fig:summaryMaps}f, we see that walls made of wood with mud take a significant proportion in the southern parts of the Western and Southern Provinces. Walls made of timber also remain prominent in the southern parts of the Western Province. Towards the center near Kigali City, we see a pattern in which sun-dried brick has become the major wall material. At the sector level in \autoref{fig:summaryMaps}j, the Western Province has more sun-dried brick in its central and northern parts, followed by wood with mud in the southern parts, with a minor proportion of timber-made walls in a few sectors. The Northern and Southern provinces share a similar pattern wherein the outer perimeter is dominated by wood-with-mud-made walls and the sun-dried bricks are concentrated towards the center near Kigali City. In Kigali City, we see that the sun-dried brick is the main wall material. For the Eastern province, its southwest and northeast parts are dominated by the sun-dried bricks, with the center concentrated with timber-made walls.

For the height distribution in \autoref{fig:summaryMaps}c,  ~\ref{fig:summaryMaps}g, and  ~\ref{fig:summaryMaps}k, we see that the dominant proportion of H:1 is slightly higher than that of H:2. In the southern part of the Southern Province, we see that H:1 is significantly higher than H:2. For the inferred macro-taxonomy distribution in  \autoref{fig:summaryMaps}d, unreinforced adobe block masonry with cement mortar (MUR+ADO, MUR+ADO+MOC) and wattle and daub (W+WWD) are the main macro-taxonomy classes, followed by unreinforced clay brick masonry with cement mortar (MUR+CL, MUR+CL+MOC). At the district and sector levels in  \autoref{fig:summaryMaps}h and  ~\ref{fig:summaryMaps}l, respectively, we see the W+WWD are more prominent in the southern part of the Southern Province and the center of the Eastern Province. For the northern part of the Eastern and Southern Province and the southern part of the Northern Province, we see that MUR+ADO and MUR+ADO+MOC are the main macro-taxonomy classes.

\begin{figure*}[t]
    \centering
    \makebox[\textwidth][c]{\includegraphics[width=7.3in]{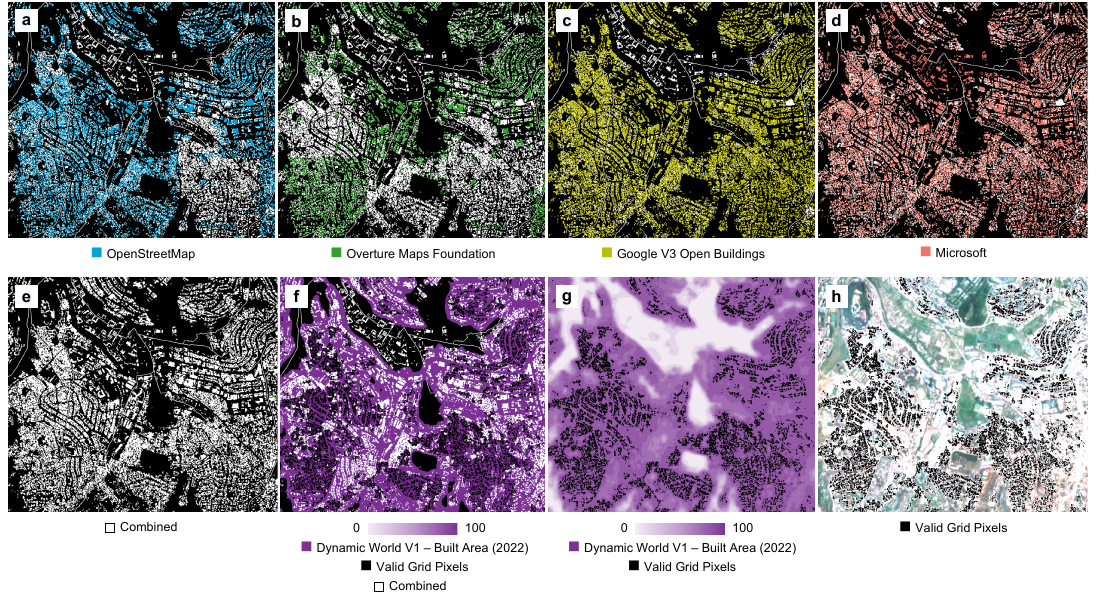}}%
    \caption{Comparison of built area information. \textbf{(a-e)} various imperfect building footprint sources and \textbf{(f-g)} the use of Dynamic World V1 near-real-time land use land cover as additional information for \textbf{(h)} possible building locations for subsequent clustering implementation of \textsc{DeepC4}.}
    \label{fig:bldgMap}
\end{figure*}

The spatial distribution of our resulting Rwandan urban morphology relies on the use of many imperfect representations of the building footprint, as shown in \autoref{fig:bldgMap}a to ~\ref{fig:bldgMap}d. In the provided example in \autoref{fig:bldgMap}, OpenStreetMap and Overture did not have building footprints in the southeastern part of the illustration. Google V3 was also not able to map the building footprints in the northern part. Among them, Microsoft seems to provide most of the building footprints combined from all four sources (see \autoref{fig:bldgMap}e). Considering these four imperfect sources to be equally imperfect due to their lack of temporal definition, the probability values from the Dynamic World V1 Built Area validated and finalized a set of building footprint pixels (see Figure \autoref{fig:bldgMap}f and  ~\ref{fig:bldgMap}g) which also closely resembles the Sentinel-2 RGB imagery in  \autoref{fig:bldgMap}h.

\subsection{Limitations}
\label{limitations}

The proposed \textsc{DeepC4} framework and the demonstration using Rwandan urban morphology are subject to several spatiotemporal limitations. The \textsc{DeepC4} model is trained on building typology data derived from the 2022 population and housing census and the 2015 dataset on building typology for Kigali City, Rwanda, and applied to 2022 satellite observations, introducing a temporal gap during which urban morphology may have changed substantially, particularly in rapidly developing peri-urban areas. The supervising information is further limited by the geographic bias inherent in using data from Kigali City only, which may reduce generalizability to rural areas and other urban contexts with different morphological characteristics. The weak supervision and label ambiguity arising from overlapping and conditionally valid class assignments introduce difficulty in evaluating predictive performance, which is further compounded by class imbalance across the joint multitask setting of roof, wall, and height prediction. These challenges present an opportunity for future fine-tuning improvement when large-scale survey data from additional regions or more recent census information becomes available.

Regarding application boundaries, despite contributing to the understanding of urban morphology in terms of spatial patterns, density, and numerical metrics, the presented indicators are limited to physical building-level attributes and do not capture functional or socioeconomic dimensions of urban exposure. The constraint-based formulation also depends on the quality and temporal consistency of census statistics relative to the satellite observation window. The framework relies on existing third-party built area products with inherent model uncertainties. The cluster-level constraints also operate at the sector level rather than at the building level. Most importantly, the cross-validation is limited to 20 sectors, which can be further improved to establish the robustness and generalizability by considering a wider and geographically diverse validation set.

\section{Conclusion and Future Work}
\label{Conclusion}

Our proposed novel \textsc{DeepC4}, a deep constrained clustering model, advances the existing classical spatial disaggregation techniques using deep learning and EO data to understand the distribution of various urban morphology characteristics at large scales. Using cluster-level constraints, multitask optimization, and multilabel conditionality, we specifically demonstrated a deep learning model that can uniquely cope with the challenges of weak supervision when inferring the Rwandan urban morphology, in terms of indicators often used in the practice of disaster risk assessment. By considering the explicit importance of well-validated census and experts' belief systems, our approach also contributes to the overall consistency of resulting deep learning-based maps, thereby improving their relative explainability and interpretability with respect to the current understanding of end users. Using several built environment data and partially available building-level groundtruth, our work has also significantly contributed new insights into the existing Rwandan exposure and physical vulnerability of GEM and METEOR towards a spatial auditing of existing coarse-grained derived information. 

For future work, we recommend extending this study with temporal variables to understand the evolution of urban morphology. We also recommend investigating other regions with higher diversity and uniformity across labels to assess the training quality and sensitivity.

\section*{Data and Code Availability}
\label{dataAndCode}
The 9.2 GB data and code are respectively available in our publicly available Zenodo (\url{https://doi.org/10.5281/zenodo.13119552}) \citep{dimasakaZenodoDeepC4} and GitHub repositories (\url{https://github.com/riskaudit/DeepC4}) \citep{dimasakaGitHubDeepC4}.
\section*{Acknowledgment}
\label{acknowledgment}
This work is funded by the UKRI Centre for Doctoral Training in Application of Artificial Intelligence to the study of Environmental Risks (AI4ER) (EP/S022961/1) and the Helmholtz Information \& Data Science Academy (HIDA) for providing financial support, enabling a short-term research stay at the Earth Observation Center in the German Aerospace Center to develop the \textsc{DeepC4}. We also acknowledge the GEM Foundation, through Andrés Abarca and Vitor Silva, for the subset of the Global Exposure Model (v2023.1.0) for research purposes. We also appreciate the initial guidance and correspondence from Charles Huyck of the METEOR project and Nicole Paul of the Uniform Africa Exposure Dataset at the outset of this research project.

\bibliographystyle{elsarticle-harv} 
\bibliography{bibliography}

\begin{thebibliography}{51}
\expandafter\ifx\csname natexlab\endcsname\relax\def\natexlab#1{#1}\fi
\providecommand{\url}[1]{\texttt{#1}}
\providecommand{\href}[2]{#2}
\providecommand{\path}[1]{#1}
\providecommand{\DOIprefix}{doi:}
\providecommand{\ArXivprefix}{arXiv:}
\providecommand{\URLprefix}{URL: }
\providecommand{\Pubmedprefix}{pmid:}
\providecommand{\doi}[1]{\href{http://dx.doi.org/#1}{\path{#1}}}
\providecommand{\Pubmed}[1]{\href{pmid:#1}{\path{#1}}}
\providecommand{\bibinfo}[2]{#2}
\ifx\xfnm\relax \def\xfnm[#1]{\unskip,\space#1}\fi
\bibitem[{Bachofer et~al.(2019)Bachofer, Braun, Adamietz, Murray, d’Angelo, Kyazze, Mumuhire and Bower}]{bachofer2019building}
\bibinfo{author}{Bachofer, F.}, \bibinfo{author}{Braun, A.}, \bibinfo{author}{Adamietz, F.}, \bibinfo{author}{Murray, S.}, \bibinfo{author}{d’Angelo, P.}, \bibinfo{author}{Kyazze, E.}, \bibinfo{author}{Mumuhire, A.P.}, \bibinfo{author}{Bower, J.}, \bibinfo{year}{2019}.
\newblock \bibinfo{title}{Building stock and building typology of kigali, rwanda}.
\newblock \bibinfo{journal}{Data} \bibinfo{volume}{4}, \bibinfo{pages}{105}.
\bibitem[{Blei et~al.(2018)Blei, Angel, Civco, Liu and Zhang}]{blei2018accuracy}
\bibinfo{author}{Blei, A.M.}, \bibinfo{author}{Angel, S.}, \bibinfo{author}{Civco, D.L.}, \bibinfo{author}{Liu, Y.}, \bibinfo{author}{Zhang, X.}, \bibinfo{year}{2018}.
\newblock \bibinfo{title}{Accuracy assessment and map comparisons for monitoring urban expansion: the atlas of urban expansion and the global human settlement layer}.
\newblock \bibinfo{journal}{Lincoln Institute of Land Policy: Cambridge, MA, USA} .
\bibitem[{Bradley et~al.(2000)Bradley, Bennett and Demiriz}]{bradley2000constrained}
\bibinfo{author}{Bradley, P.S.}, \bibinfo{author}{Bennett, K.P.}, \bibinfo{author}{Demiriz, A.}, \bibinfo{year}{2000}.
\newblock \bibinfo{title}{Constrained k-means clustering}.
\newblock \bibinfo{journal}{Microsoft Research, Redmond} \bibinfo{volume}{20}, \bibinfo{pages}{0}.
\bibitem[{Braun et~al.(2019)Braun, Warth, Bachofer and Hochschild}]{braun2019}
\bibinfo{author}{Braun, A.}, \bibinfo{author}{Warth, G.}, \bibinfo{author}{Bachofer, F.}, \bibinfo{author}{Hochschild, V.}, \bibinfo{year}{2019}.
\newblock \bibinfo{title}{Identification of roof materials in high-resolution multispectral images for urban planning and monitoring}, pp. \bibinfo{pages}{1--4}.
\newblock \DOIprefix\doi{10.1109/JURSE.2019.8809026}.
\bibitem[{Brown et~al.(2022)Brown, Brumby, Guzder-Williams, Birch, Hyde, Mazzariello, Czerwinski, Pasquarella, Haertel, Ilyushchenko et~al.}]{brown2022dynamic}
\bibinfo{author}{Brown, C.F.}, \bibinfo{author}{Brumby, S.P.}, \bibinfo{author}{Guzder-Williams, B.}, \bibinfo{author}{Birch, T.}, \bibinfo{author}{Hyde, S.B.}, \bibinfo{author}{Mazzariello, J.}, \bibinfo{author}{Czerwinski, W.}, \bibinfo{author}{Pasquarella, V.J.}, \bibinfo{author}{Haertel, R.}, \bibinfo{author}{Ilyushchenko, S.}, et~al., \bibinfo{year}{2022}.
\newblock \bibinfo{title}{Dynamic world, near real-time global 10 m land use land cover mapping}.
\newblock \bibinfo{journal}{Scientific Data} \bibinfo{volume}{9}, \bibinfo{pages}{251}.
\bibitem[{Brzev et~al.(2013)Brzev, Scawthorn, Charleson, Allen, Greene, Jaiswal and Silva}]{brzev2013gem}
\bibinfo{author}{Brzev, S.}, \bibinfo{author}{Scawthorn, C.}, \bibinfo{author}{Charleson, A.W.}, \bibinfo{author}{Allen, L.}, \bibinfo{author}{Greene, M.}, \bibinfo{author}{Jaiswal, K.}, \bibinfo{author}{Silva, V.}, \bibinfo{year}{2013}.
\newblock \bibinfo{title}{GEM building taxonomy (Version 2.0)}.
\newblock \bibinfo{type}{Technical Report}. GEM Foundation.
\bibitem[{Clark et~al.(2023)Clark, Bedada, Huff, Lita, Newbury and Pacifici}]{clark2023investigating}
\bibinfo{author}{Clark, C.N.}, \bibinfo{author}{Bedada, A.}, \bibinfo{author}{Huff, B.}, \bibinfo{author}{Lita, B.}, \bibinfo{author}{Newbury, A.B.}, \bibinfo{author}{Pacifici, F.}, \bibinfo{year}{2023}.
\newblock \bibinfo{title}{Investigating the resolution-performance trade-off of object detection models in support of the sustainable development goals}.
\newblock \bibinfo{journal}{IEEE Journal of Selected Topics in Applied Earth Observations and Remote Sensing} \bibinfo{volume}{16}, \bibinfo{pages}{5695--5713}.
\bibitem[{{Copernicus Sentinel data}(2024a)}]{sentinel2}
\bibinfo{author}{{Copernicus Sentinel data}}, \bibinfo{year}{2024}a.
\newblock \bibinfo{title}{{Harmonized Sentinel-2 MSI: MultiSpectral Instrument, Level-2A}}.
\newblock \bibinfo{howpublished}{\url{https://developers.google.com/earth-engine/datasets/catalog/COPERNICUS_S2_SR_HARMONIZED}}.
\newblock \bibinfo{note}{Accessed: 2024-07-30}.
\bibitem[{{Copernicus Sentinel data}(2024b)}]{sentinel1}
\bibinfo{author}{{Copernicus Sentinel data}}, \bibinfo{year}{2024}b.
\newblock \bibinfo{title}{{Sentinel-1 SAR GRD: C-band Synthetic Aperture Radar Ground Range Detected, log scaling}}.
\newblock \bibinfo{howpublished}{\url{https://developers.google.com/earth-engine/datasets/catalog/COPERNICUS_S1_GRD}}.
\newblock \bibinfo{note}{Accessed: 2024-07-30}.
\bibitem[{{Copernicus Sentinel data}(2024c)}]{sentinel2cloud}
\bibinfo{author}{{Copernicus Sentinel data}}, \bibinfo{year}{2024}c.
\newblock \bibinfo{title}{{Sentinel-2: Cloud Probability}}.
\newblock \bibinfo{howpublished}{\url{https://developers.google.com/earth-engine/datasets/catalog/COPERNICUS_S2_CLOUD_PROBABILITY}}.
\newblock \bibinfo{note}{Accessed: 2024-07-30}.
\bibitem[{Darin et~al.(2022)Darin, Ku{\'e}pi{\'e}, Bassinga, Boo, Tatem and Reeve}]{darin2022population}
\bibinfo{author}{Darin, E.}, \bibinfo{author}{Ku{\'e}pi{\'e}, M.}, \bibinfo{author}{Bassinga, H.}, \bibinfo{author}{Boo, G.}, \bibinfo{author}{Tatem, A.J.}, \bibinfo{author}{Reeve, P.}, \bibinfo{year}{2022}.
\newblock \bibinfo{title}{The population seen from space: when satellite images come to the rescue of the census}.
\newblock \bibinfo{journal}{Population} \bibinfo{volume}{77}, \bibinfo{pages}{437--464}.
\bibitem[{Demiriz et~al.(2008)Demiriz, Bennett and Bradley}]{demiriz2008using}
\bibinfo{author}{Demiriz, A.}, \bibinfo{author}{Bennett, K.P.}, \bibinfo{author}{Bradley, P.S.}, \bibinfo{year}{2008}.
\newblock \bibinfo{title}{Using assignment constraints to avoid empty clusters in k-means clustering}.
\newblock \bibinfo{journal}{Constrained clustering: advances in algorithms, theory, and applications} \bibinfo{volume}{201}.
\bibitem[{Dimasaka(2024a)}]{dimasakaZenodoDeepC4}
\bibinfo{author}{Dimasaka, J.}, \bibinfo{year}{2024}a.
\newblock \bibinfo{title}{{Deep Conditional Census-Constrained Clustering (DeepC4) for Large-scale Multi-task Disaggregation of Urban Morphology}}.
\newblock \URLprefix \url{https://doi.org/10.5281/zenodo.13119552}, \DOIprefix\doi{10.5281/zenodo.13119552}.
\bibitem[{Dimasaka(2024b)}]{dimasakaGitHubDeepC4}
\bibinfo{author}{Dimasaka, J.}, \bibinfo{year}{2024}b.
\newblock \bibinfo{title}{{DeepC4}}.
\newblock \URLprefix \url{https://github.com/riskaudit/DeepC4}.
\bibitem[{Gei{\ss} et~al.(2023)Gei{\ss}, Priesmeier, Aravena~Pelizari, Soto~Calderon, Schoepfer, Riedlinger, Villar~Vega, Santa~Mar{\'\i}a, G{\'o}mez~Zapata, Pittore et~al.}]{geiss2023benefits}
\bibinfo{author}{Gei{\ss}, C.}, \bibinfo{author}{Priesmeier, P.}, \bibinfo{author}{Aravena~Pelizari, P.}, \bibinfo{author}{Soto~Calderon, A.R.}, \bibinfo{author}{Schoepfer, E.}, \bibinfo{author}{Riedlinger, T.}, \bibinfo{author}{Villar~Vega, M.}, \bibinfo{author}{Santa~Mar{\'\i}a, H.}, \bibinfo{author}{G{\'o}mez~Zapata, J.C.}, \bibinfo{author}{Pittore, M.}, et~al., \bibinfo{year}{2023}.
\newblock \bibinfo{title}{Benefits of global earth observation missions for disaggregation of exposure data and earthquake loss modeling: evidence from santiago de chile}.
\newblock \bibinfo{journal}{Natural Hazards} \bibinfo{volume}{119}, \bibinfo{pages}{779--804}.
\bibitem[{Gei{\ss} et~al.(2018)Gei{\ss}, Thoma and Taubenb{\"o}ck}]{geiss2018cost}
\bibinfo{author}{Gei{\ss}, C.}, \bibinfo{author}{Thoma, M.}, \bibinfo{author}{Taubenb{\"o}ck, H.}, \bibinfo{year}{2018}.
\newblock \bibinfo{title}{Cost-sensitive multitask active learning for characterization of urban environments with remote sensing}.
\newblock \bibinfo{journal}{IEEE Geoscience and Remote Sensing Letters} \bibinfo{volume}{15}, \bibinfo{pages}{922--926}.
\bibitem[{{Geofabrik GmbH} and {OpenStreetMap Contributors}(2018)}]{geofabrikOSM}
\bibinfo{author}{{Geofabrik GmbH}}, \bibinfo{author}{{OpenStreetMap Contributors}}, \bibinfo{year}{2018}.
\newblock \bibinfo{title}{{OpenStreetMap Data Extracts}}.
\newblock \bibinfo{note}{URL \url{https://download.geofabrik.de/}. Accessed: 2024-07-30}.
\bibitem[{Gevaert et~al.(2021)Gevaert, Carman, Rosman, Georgiadou and Soden}]{gevaert2021fairness}
\bibinfo{author}{Gevaert, C.M.}, \bibinfo{author}{Carman, M.}, \bibinfo{author}{Rosman, B.}, \bibinfo{author}{Georgiadou, Y.}, \bibinfo{author}{Soden, R.}, \bibinfo{year}{2021}.
\newblock \bibinfo{title}{Fairness and accountability of ai in disaster risk management: Opportunities and challenges}.
\newblock \bibinfo{journal}{Patterns} \bibinfo{volume}{2}.
\bibitem[{{Global Administrative Areas}(2022)}]{gadm}
\bibinfo{author}{{Global Administrative Areas}}, \bibinfo{year}{2022}.
\newblock \bibinfo{title}{{GADM} database of global administrative areas, version 4.1}.
\newblock \bibinfo{howpublished}{\url{https://gadm.org/maps/NOR.html}}.
\newblock \bibinfo{note}{To view a copy of the license of this work, visit \url{https://gadm.org/license.html}. Accessed: 2024-07-30.}
\bibitem[{G{\'o}mez~Zapata et~al.(2022)G{\'o}mez~Zapata, Pittore, Cotton, Lilienkamp, Shinde, Aguirre and Santa~Mar{\'\i}a}]{gomez2022epistemic}
\bibinfo{author}{G{\'o}mez~Zapata, J.C.}, \bibinfo{author}{Pittore, M.}, \bibinfo{author}{Cotton, F.}, \bibinfo{author}{Lilienkamp, H.}, \bibinfo{author}{Shinde, S.}, \bibinfo{author}{Aguirre, P.}, \bibinfo{author}{Santa~Mar{\'\i}a, H.}, \bibinfo{year}{2022}.
\newblock \bibinfo{title}{Epistemic uncertainty of probabilistic building exposure compositions in scenario-based earthquake loss models}.
\newblock \bibinfo{journal}{Bulletin of Earthquake Engineering} \bibinfo{volume}{20}, \bibinfo{pages}{2401--2438}.
\bibitem[{Gorelick et~al.(2017)Gorelick, Hancher, Dixon, Ilyushchenko, Thau and Moore}]{gorelick2017google}
\bibinfo{author}{Gorelick, N.}, \bibinfo{author}{Hancher, M.}, \bibinfo{author}{Dixon, M.}, \bibinfo{author}{Ilyushchenko, S.}, \bibinfo{author}{Thau, D.}, \bibinfo{author}{Moore, R.}, \bibinfo{year}{2017}.
\newblock \bibinfo{title}{Google earth engine: Planetary-scale geospatial analysis for everyone}.
\newblock \bibinfo{journal}{Remote Sensing of Environment} \URLprefix \url{https://doi.org/10.1016/j.rse.2017.06.031}, \DOIprefix\doi{10.1016/j.rse.2017.06.031}.
\bibitem[{Gouveia et~al.(2024)Gouveia, Silva, Lopes, Moreira, Torres and Simas~Guerreiro}]{gouveia2024automated}
\bibinfo{author}{Gouveia, F.}, \bibinfo{author}{Silva, V.}, \bibinfo{author}{Lopes, J.}, \bibinfo{author}{Moreira, R.S.}, \bibinfo{author}{Torres, J.M.}, \bibinfo{author}{Simas~Guerreiro, M.}, \bibinfo{year}{2024}.
\newblock \bibinfo{title}{Automated identification of building features with deep learning for risk analysis}.
\newblock \bibinfo{journal}{Discover Applied Sciences} \bibinfo{volume}{6}, \bibinfo{pages}{466}.
\bibitem[{Gr{\"u}nthal(1998)}]{grunthal1998ems}
\bibinfo{author}{Gr{\"u}nthal, G.}, \bibinfo{year}{1998}.
\newblock \bibinfo{title}{European macroseismic scale 1998} \bibinfo{volume}{15}, \bibinfo{pages}{1--97}.
\bibitem[{Huangfu and Hall(2018)}]{huangfu2018parallelizing}
\bibinfo{author}{Huangfu, Q.}, \bibinfo{author}{Hall, J.J.}, \bibinfo{year}{2018}.
\newblock \bibinfo{title}{Parallelizing the dual revised simplex method}.
\newblock \bibinfo{journal}{Mathematical Programming Computation} \bibinfo{volume}{10}, \bibinfo{pages}{119--142}.
\bibitem[{Huyck et~al.(2019)Huyck, Hu, Amyx, Esquivias, Huyck and Eguchi}]{huyck2019meteor}
\bibinfo{author}{Huyck, C.}, \bibinfo{author}{Hu, Z.}, \bibinfo{author}{Amyx, P.}, \bibinfo{author}{Esquivias, G.}, \bibinfo{author}{Huyck, M.}, \bibinfo{author}{Eguchi, M.}, \bibinfo{year}{2019}.
\newblock \bibinfo{title}{METEOR: exposure data classification, metadata population and confidence assessment. Report M3. 2/P}.
\newblock \bibinfo{type}{Technical Report} \bibinfo{number}{M3. 2/P}. {British Geological Survey}.
\bibitem[{Jaiswal et~al.(2010)Jaiswal, Wald and Porter}]{jaiswal2010global}
\bibinfo{author}{Jaiswal, K.}, \bibinfo{author}{Wald, D.}, \bibinfo{author}{Porter, K.}, \bibinfo{year}{2010}.
\newblock \bibinfo{title}{A global building inventory for earthquake loss estimation and risk management}.
\newblock \bibinfo{journal}{Earthquake Spectra} \bibinfo{volume}{26}, \bibinfo{pages}{731--748}.
\bibitem[{Kingma(2014)}]{kingma2014adam}
\bibinfo{author}{Kingma, D.P.}, \bibinfo{year}{2014}.
\newblock \bibinfo{title}{Adam: A method for stochastic optimization}.
\newblock \bibinfo{journal}{arXiv preprint arXiv:1412.6980} .
\bibitem[{Kircher et~al.(2006)Kircher, Whitman and Holmes}]{kircher2006hazus}
\bibinfo{author}{Kircher, C.A.}, \bibinfo{author}{Whitman, R.V.}, \bibinfo{author}{Holmes, W.T.}, \bibinfo{year}{2006}.
\newblock \bibinfo{title}{Hazus earthquake loss estimation methods}.
\newblock \bibinfo{journal}{Natural Hazards Review} \bibinfo{volume}{7}, \bibinfo{pages}{45--59}.
\bibitem[{Koppel et~al.(2017)Koppel, Zalite, Voormansik and Jagdhuber}]{koppel2017}
\bibinfo{author}{Koppel, K.}, \bibinfo{author}{Zalite, K.}, \bibinfo{author}{Voormansik, K.}, \bibinfo{author}{Jagdhuber, T.}, \bibinfo{year}{2017}.
\newblock \bibinfo{title}{Sensitivity of sentinel-1 backscatter to characteristics of buildings}.
\newblock \bibinfo{journal}{International Journal of Remote Sensing} \bibinfo{volume}{38}, \bibinfo{pages}{6298–6318}.
\newblock \DOIprefix\doi{10.1080/01431161.2017.1353160}.
\bibitem[{Labetski et~al.(2023)Labetski, Vitalis, Biljecki, Arroyo~Ohori and Stoter}]{labetski20233d}
\bibinfo{author}{Labetski, A.}, \bibinfo{author}{Vitalis, S.}, \bibinfo{author}{Biljecki, F.}, \bibinfo{author}{Arroyo~Ohori, K.}, \bibinfo{author}{Stoter, J.}, \bibinfo{year}{2023}.
\newblock \bibinfo{title}{3d building metrics for urban morphology}.
\newblock \bibinfo{journal}{International Journal of Geographical Information Science} \bibinfo{volume}{37}, \bibinfo{pages}{36--67}.
\bibitem[{Lafabr{\`e}gue and Gan{\c{c}}arski(2025)}]{lafabregue2025samarah}
\bibinfo{author}{Lafabr{\`e}gue, B.}, \bibinfo{author}{Gan{\c{c}}arski, P.}, \bibinfo{year}{2025}.
\newblock \bibinfo{title}{I-samarah, an incremental constrained clustering applied to remote sensing images}.
\newblock \bibinfo{journal}{Neural Computing and Applications} , \bibinfo{pages}{1--25}.
\bibitem[{Lampert et~al.(2019)Lampert, Lafabregue, Serrette, Vrain, Gan{\c{c}}arski et~al.}]{lampert2019constrained}
\bibinfo{author}{Lampert, T.}, \bibinfo{author}{Lafabregue, B.}, \bibinfo{author}{Serrette, N.}, \bibinfo{author}{Vrain, C.}, \bibinfo{author}{Gan{\c{c}}arski, P.}, et~al., \bibinfo{year}{2019}.
\newblock \bibinfo{title}{Constrained distance-based clustering for satellite image time-series}.
\newblock \bibinfo{journal}{IEEE Journal of Selected Topics in Applied Earth Observations and Remote Sensing} \bibinfo{volume}{12}, \bibinfo{pages}{4606--4621}.
\bibitem[{Meinel et~al.(2009)Meinel, Hecht and Herold}]{meinel2009analyzing}
\bibinfo{author}{Meinel, G.}, \bibinfo{author}{Hecht, R.}, \bibinfo{author}{Herold, H.}, \bibinfo{year}{2009}.
\newblock \bibinfo{title}{Analyzing building stock using topographic maps and gis}.
\newblock \bibinfo{journal}{Building Research \& Information} \bibinfo{volume}{37}, \bibinfo{pages}{468--482}.
\bibitem[{{Microsoft}(2024)}]{microsoft}
\bibinfo{author}{{Microsoft}}, \bibinfo{year}{2024}.
\newblock \bibinfo{title}{{Global ML Building Footprints}}.
\newblock \bibinfo{note}{URL: \url{https://github.com/microsoft/GlobalMLBuildingFootprints}. Accessed: 2024-07-30}.
\bibitem[{{National Institute of Statistics Rwanda}(2023)}]{nisr}
\bibinfo{author}{{National Institute of Statistics Rwanda}}, \bibinfo{year}{2023}.
\newblock \bibinfo{title}{{Fifth Population and Housing Census - 2022}}.
\newblock \bibinfo{howpublished}{\url{https://www.statistics.gov.rw/datasource/fifth-population-and-housing-census-2022}}.
\newblock \bibinfo{note}{This work is licensed under the Norwegian Licence for Creative Commons Attribution 4.0 International License. To view a copy of this license, visit \url{https://creativecommons.org/licenses/by/4.0/}. Accessed: 2024-07-30}.
\bibitem[{{Overture Maps Foundation}(2023)}]{overture}
\bibinfo{author}{{Overture Maps Foundation}}, \bibinfo{year}{2023}.
\newblock \bibinfo{title}{{Overture Foundation Building Footprints}}.
\newblock \bibinfo{note}{URL: \url{https://beta.source.coop/repositories/cholmes/overture}. Accessed: 2024-07-30}.
\bibitem[{Paul et~al.(2022)Paul, Silva and Amo-Oduro}]{paul2022development}
\bibinfo{author}{Paul, N.}, \bibinfo{author}{Silva, V.}, \bibinfo{author}{Amo-Oduro, D.}, \bibinfo{year}{2022}.
\newblock \bibinfo{title}{Development of a uniform exposure model for the african continent for use in disaster risk assessment}.
\newblock \bibinfo{journal}{International Journal of Disaster Risk Reduction} \bibinfo{volume}{71}, \bibinfo{pages}{102823}.
\bibitem[{Pesaresi et~al.(2024)Pesaresi, Schiavina, Politis, Freire, Krasnodebska, Uhl, Carioli, Corbane, Dijkstra, Florio et~al.}]{pesaresi2024advances}
\bibinfo{author}{Pesaresi, M.}, \bibinfo{author}{Schiavina, M.}, \bibinfo{author}{Politis, P.}, \bibinfo{author}{Freire, S.}, \bibinfo{author}{Krasnodebska, K.}, \bibinfo{author}{Uhl, J.H.}, \bibinfo{author}{Carioli, A.}, \bibinfo{author}{Corbane, C.}, \bibinfo{author}{Dijkstra, L.}, \bibinfo{author}{Florio, P.}, et~al., \bibinfo{year}{2024}.
\newblock \bibinfo{title}{Advances on the global human settlement layer by joint assessment of earth observation and population survey data}.
\newblock \bibinfo{journal}{International Journal of Digital Earth} \bibinfo{volume}{17}, \bibinfo{pages}{2390454}.
\bibitem[{Petrarulo et~al.(2022)Petrarulo, Lyon, Paudyal, Zakaria and Sokile}]{petrarulo2022meteor}
\bibinfo{author}{Petrarulo, L.}, \bibinfo{author}{Lyon, A.}, \bibinfo{author}{Paudyal, A.}, \bibinfo{author}{Zakaria, S.}, \bibinfo{author}{Sokile, C.}, \bibinfo{year}{2022}.
\newblock \bibinfo{title}{METEOR: Legacy Evaluation Report M2. 11/P}.
\newblock \bibinfo{type}{Technical Report} \bibinfo{number}{M2. 11/P}. {British Geological Survey}.
\bibitem[{Pittore et~al.(2018)Pittore, Haas and Megalooikonomou}]{pittore2018risk}
\bibinfo{author}{Pittore, M.}, \bibinfo{author}{Haas, M.}, \bibinfo{author}{Megalooikonomou, K.G.}, \bibinfo{year}{2018}.
\newblock \bibinfo{title}{Risk-oriented, bottom-up modeling of building portfolios with faceted taxonomies}.
\newblock \bibinfo{journal}{Frontiers in Built Environment} \bibinfo{volume}{4}, \bibinfo{pages}{41}.
\bibitem[{Seljom et~al.(2021)Seljom, Kvalbein, Hellemo, Kaut and Ortiz}]{seljom2021stochastic}
\bibinfo{author}{Seljom, P.}, \bibinfo{author}{Kvalbein, L.}, \bibinfo{author}{Hellemo, L.}, \bibinfo{author}{Kaut, M.}, \bibinfo{author}{Ortiz, M.M.}, \bibinfo{year}{2021}.
\newblock \bibinfo{title}{Stochastic modelling of variable renewables in long-term energy models: Dataset, scenario generation \& quality of results}.
\newblock \bibinfo{journal}{Energy} \bibinfo{volume}{236}, \bibinfo{pages}{121415}.
\bibitem[{Silva et~al.(2024)Silva, Sousa, Ribeiro~Gouveia, Lopes and Guerreiro}]{silva2024building}
\bibinfo{author}{Silva, V.}, \bibinfo{author}{Sousa, R.}, \bibinfo{author}{Ribeiro~Gouveia, F.}, \bibinfo{author}{Lopes, J.}, \bibinfo{author}{Guerreiro, M.J.}, \bibinfo{year}{2024}.
\newblock \bibinfo{title}{A building imagery database for the calibration of machine learning algorithms}.
\newblock \bibinfo{journal}{Earthquake Spectra} \bibinfo{volume}{40}, \bibinfo{pages}{1577--1590}.
\bibitem[{Sirko et~al.(2021)Sirko, Kashubin, Ritter, Annkah, Bouchareb, Dauphin, Keysers, Neumann, Cisse and Quinn}]{sirko2021continental}
\bibinfo{author}{Sirko, W.}, \bibinfo{author}{Kashubin, S.}, \bibinfo{author}{Ritter, M.}, \bibinfo{author}{Annkah, A.}, \bibinfo{author}{Bouchareb, Y.S.E.}, \bibinfo{author}{Dauphin, Y.}, \bibinfo{author}{Keysers, D.}, \bibinfo{author}{Neumann, M.}, \bibinfo{author}{Cisse, M.}, \bibinfo{author}{Quinn, J.}, \bibinfo{year}{2021}.
\newblock \bibinfo{title}{Continental-scale building detection from high resolution satellite imagery}.
\newblock \bibinfo{journal}{arXiv preprint arXiv:2107.12283} \bibinfo{note}{URL: \url{https://beta.source.coop/repositories/cholmes/google-open-buildings/description/}. Accessed: 2024-07-30}.
\bibitem[{Stevens et~al.(2015)Stevens, Gaughan, Linard and Tatem}]{stevens2015disaggregating}
\bibinfo{author}{Stevens, F.R.}, \bibinfo{author}{Gaughan, A.E.}, \bibinfo{author}{Linard, C.}, \bibinfo{author}{Tatem, A.J.}, \bibinfo{year}{2015}.
\newblock \bibinfo{title}{Disaggregating census data for population mapping using random forests with remotely-sensed and ancillary data}.
\newblock \bibinfo{journal}{PloS one} \bibinfo{volume}{10}, \bibinfo{pages}{e0107042}.
\bibitem[{Tatem(2017)}]{tatem2017worldpop}
\bibinfo{author}{Tatem, A.J.}, \bibinfo{year}{2017}.
\newblock \bibinfo{title}{Worldpop, open data for spatial demography}.
\newblock \bibinfo{journal}{Scientific data} \bibinfo{volume}{4}, \bibinfo{pages}{1--4}.
\bibitem[{{UN}(2015)}]{un2015sdg}
\bibinfo{author}{{UN}}, \bibinfo{year}{2015}.
\newblock \bibinfo{title}{Resolution A/RES/70/1. Transforming Our World, the 2030 Agenda for Sustainable Development}.
\newblock \bibinfo{type}{Technical Report}. {United Nations}.
\bibitem[{{United Nations (UN)}(2015)}]{un2015sfdrr}
\bibinfo{author}{{United Nations (UN)}}, \bibinfo{year}{2015}.
\newblock \bibinfo{title}{Sendai Framework for Disaster Risk Reduction 2015–2030}.
\newblock \bibinfo{type}{Technical Report}. {United Nations}.
\bibitem[{Wagstaff et~al.(2001)Wagstaff, Cardie, Rogers, Schr{\"o}dl et~al.}]{wagstaff2001constrained}
\bibinfo{author}{Wagstaff, K.}, \bibinfo{author}{Cardie, C.}, \bibinfo{author}{Rogers, S.}, \bibinfo{author}{Schr{\"o}dl, S.}, et~al., \bibinfo{year}{2001}.
\newblock \bibinfo{title}{Constrained k-means clustering with background knowledge}, in: \bibinfo{booktitle}{Icml}, pp. \bibinfo{pages}{577--584}.
\bibitem[{Wardrop et~al.(2018)Wardrop, Jochem, Bird, Chamberlain, Clarke, Kerr, Bengtsson, Juran, Seaman and Tatem}]{wardrop2018spatially}
\bibinfo{author}{Wardrop, N.A.}, \bibinfo{author}{Jochem, W.C.}, \bibinfo{author}{Bird, T.J.}, \bibinfo{author}{Chamberlain, H.R.}, \bibinfo{author}{Clarke, D.}, \bibinfo{author}{Kerr, D.}, \bibinfo{author}{Bengtsson, L.}, \bibinfo{author}{Juran, S.}, \bibinfo{author}{Seaman, V.}, \bibinfo{author}{Tatem, A.J.}, \bibinfo{year}{2018}.
\newblock \bibinfo{title}{Spatially disaggregated population estimates in the absence of national population and housing census data}.
\newblock \bibinfo{journal}{Proceedings of the National Academy of Sciences} \bibinfo{volume}{115}, \bibinfo{pages}{3529--3537}.
\bibitem[{Zhou et~al.(2023)Zhou, Huang, Scheuer, Wang, Zhou and Liu}]{zhou2023high}
\bibinfo{author}{Zhou, X.}, \bibinfo{author}{Huang, Z.}, \bibinfo{author}{Scheuer, B.}, \bibinfo{author}{Wang, H.}, \bibinfo{author}{Zhou, G.}, \bibinfo{author}{Liu, Y.}, \bibinfo{year}{2023}.
\newblock \bibinfo{title}{High-resolution estimation of building energy consumption at the city level}.
\newblock \bibinfo{journal}{Energy} \bibinfo{volume}{275}, \bibinfo{pages}{127476}.
\bibitem[{Zhu et~al.(2010)Zhu, Wang and Li}]{zhu2010data}
\bibinfo{author}{Zhu, S.}, \bibinfo{author}{Wang, D.}, \bibinfo{author}{Li, T.}, \bibinfo{year}{2010}.
\newblock \bibinfo{title}{Data clustering with size constraints}.
\newblock \bibinfo{journal}{Knowledge-Based Systems} \bibinfo{volume}{23}, \bibinfo{pages}{883--889}.

\end{thebibliography}
\newpage
\appendix
\section{Discrete conditional probability values among urban morphology indicators}
\label{appendix:conditionalProb}
\setcounter{table}{0}

\begin{table}[htpb!]
\centering
\caption{Relationship between wall and macro-taxonomy classes derived from \citet{paul2022development}.}
\label{appendixtab:wall2macro}
\resizebox{\columnwidth}{!}{%
\begin{tabular}{llll}
\hline
\textbf{Wall} & \textbf{Type} & \textbf{Macro-Taxonomy} & \textbf{Probability} \\ \hline
Wood with mud & Urban & W+WWD/LWAL & 1 \\
Sun-dried bricks & Urban & MUR+ADO+MOC/LWAL & 0.3 \\
Sun-dried bricks & Urban & MUR+CL+MOC/LWAL & 0.2 \\
Sun-dried bricks & Urban & MUR+ADO/LWAL & 0.3 \\
Sun-dried bricks & Urban & MUR+CL/LWAL & 0.2 \\
Others & Urban & MATO & 1 \\
Cement blocks & Urban & MUR+CB/LWAL & 0.8 \\
Cement blocks & Urban & MCF+CB/LWAL & 0.2 \\
Concrete & Urban & CR/LFINF & 0.8 \\
Concrete & Urban & CR/LWAL & 0.2 \\
Stone & Urban & MUR+STDRE+MOC/LWAL & 0.5 \\
Stone & Urban & MUR+STDRE/LWAL & 0.5 \\
Timber & Urban & W/LWAL & 1 \\
Burnt bricks & Urban & CR/LFINF & 0.25 \\
Burnt bricks & Urban & MUR+CL+MOC/LWAL & 0.25 \\
Burnt bricks & Urban & MUR+CL/LWAL & 0.412 \\
Burnt bricks & Urban & MCF+CL/LWAL & 0.088 \\ \hline
Wood with mud & Rural & W+WWD/LWAL & 1 \\
Sun-dried bricks & Rural & MUR+ADO+MOC/LWAL & 0.4 \\
Sun-dried bricks & Rural & MUR+CL+MOC/LWAL & 0.1 \\
Sun-dried bricks & Rural & MUR+ADO/LWAL & 0.4 \\
Sun-dried bricks & Rural & MUR+CL/LWAL & 0.1 \\
Others & Rural & MATO & 1 \\
Cement blocks & Rural & MUR+CB/LWAL & 1 \\
Concrete & Rural & CR/LFINF & 0.5 \\
Concrete & Rural & CR/LWAL & 0.5 \\
Stone & Rural & MUR+STRUB+MOC/LWAL & 0.5 \\
Stone & Rural & MUR+STRUB/LWAL & 0.5 \\
Timber & Rural & W/LWAL & 1 \\
Burnt bricks & Rural & MUR+CL+MOC/LWAL & 0.5 \\
Burnt bricks & Rural & MUR+CL/LWAL & 0.5 \\ \hline
\end{tabular}%
}
\end{table}

\begin{table}[t!]
\centering
\caption{Relationship between macro-taxonomy and height classes \citep{paul2022development}.}
\label{macro2height}
\resizebox{\columnwidth}{!}{%
\begin{tabular}{lll}
\hline
\textbf{Macro-Taxonomy} & \textbf{Height} & \textbf{Probability} \\ \hline
CR/LFINF & H:1 & 0.1 \\
CR/LFINF & H:2 & 0.35 \\
CR/LFINF & H:3 & 0.35 \\
CR/LFINF & HBET:4-7 & 0.15 \\
CR/LFINF & HBET:8+ & 0.05 \\ \hline
CR/LWAL & H:1 & 0.1 \\
CR/LWAL & H:2 & 0.35 \\
CR/LWAL & H:3 & 0.35 \\
CR/LWAL & HBET:4-7 & 0.15 \\
CR/LWAL & HBET:8+ & 0.05 \\ \hline
MATO & H:1 & 1 \\ \hline
MCF+CB/LWAL & H:1 & 0.45 \\
MCF+CB/LWAL & H:2 & 0.45 \\
MCF+CB/LWAL & HBET:3-6 & 0.1 \\ \hline
MCF+CL/LWAL & H:1 & 0.45 \\
MCF+CL/LWAL & H:2 & 0.45 \\
MCF+CL/LWAL & HBET:3-6 & 0.1 \\ \hline
MUR+ADO+MOC/LWAL & H:1 & 0.475 \\
MUR+ADO+MOC/LWAL & H:2 & 0.475 \\
MUR+ADO+MOC/LWAL & HBET:3-6 & 0.05 \\ \hline
MUR+ADO/LWAL & H:1 & 0.475 \\
MUR+ADO/LWAL & H:2 & 0.475 \\
MUR+ADO/LWAL & HBET:3-6 & 0.05 \\ \hline
MUR+CB/LWAL & H:1 & 0.475 \\
MUR+CB/LWAL & H:2 & 0.475 \\
MUR+CB/LWAL & HBET:3-6 & 0.05 \\ \hline
MUR+CL+MOC/LWAL & H:1 & 0.475 \\
MUR+CL+MOC/LWAL & H:2 & 0.475 \\
MUR+CL+MOC/LWAL & HBET:3-6 & 0.05 \\ \hline
MUR+CL/LWAL & H:1 & 0.475 \\
MUR+CL/LWAL & H:2 & 0.475 \\
MUR+CL/LWAL & HBET:3-6 & 0.05 \\ \hline
MUR+STDRE+MOC/LWAL & H:1 & 0.475 \\
MUR+STDRE+MOC/LWAL & H:2 & 0.475 \\
MUR+STDRE+MOC/LWAL & HBET:3-6 & 0.05 \\ \hline
MUR+STDRE/LWAL & H:1 & 0.475 \\
MUR+STDRE/LWAL & H:2 & 0.475 \\
MUR+STDRE/LWAL & HBET:3-6 & 0.05 \\ \hline
MUR+STRUB+MOC/LWAL & H:1 & 0.475 \\
MUR+STRUB+MOC/LWAL & H:2 & 0.475 \\
MUR+STRUB+MOC/LWAL & HBET:3-6 & 0.05 \\ \hline
MUR+STRUB/LWAL & H:1 & 0.475 \\
MUR+STRUB/LWAL & H:2 & 0.475 \\
MUR+STRUB/LWAL & HBET:3-6 & 0.05 \\ \hline
W+WWD/LWAL & H:1 & 0.7 \\
W+WWD/LWAL & H:2 & 0.3 \\ \hline
W/LWAL & H:1 & 0.4 \\
W/LWAL & H:2 & 0.4 \\
W/LWAL & H:3 & 0.2 \\ \hline
\end{tabular}%
}
\end{table}

\begin{table}[htpb!]
\centering
\caption{Relationship between height classes and dwellings \citep{paul2022development}.}
\label{height2dwellings}
\resizebox{\columnwidth}{!}{%
\begin{tabular}{llll}
\hline
\textbf{Height} & \textbf{Type} & \textbf{Probability} & \textbf{Dwellings} \\ \hline
H:1 & Dwelling-house & 1 & 1 \\ \hline
H:2 & Dwelling-house & 0.95 & 1 \\
H:2 & Townhouse & 0.05 & 2 \\ \hline
H:3 & Dwelling-house & 0.7 & 1 \\
H:3 & Townhouse & 0.2 & 3 \\
H:3 & Apartment/Flat & 0.1 & 8 \\ \hline
HBET:3-6 & Townhouse & 0.25 & 4 \\
HBET:3-6 & Apartment/Flat & 0.75 & 12 \\ \hline
HBET:4-7 & Apartment/Flat & 1 & 12 \\ \hline
HBET:8+ & Apartment/Flat & 1 & 16 \\ \hline
\end{tabular}%
}
\end{table}

\clearpage
\section{Mapping scheme between the target urban morphology indicators and the classification of 2015 Rwanda building groundtruth information \citep{bachofer2019building}.}
\label{appendix:mappingscheme}
\setcounter{table}{0}

\begin{table}[htpb!]
\centering
\caption{Assumed relationship between available building type labels ($Y_{C^{h}}$) and discrete target census roof classes ($C^{h}_{s_i}$).}
\label{tab:my-table}
\resizebox{\columnwidth}{!}{%
\begin{tabular}{lcccc}
\hline \\[-1em]
\multirow{2}{*}{\textbf{Building Type Labels}} & \multicolumn{4}{c}{\textbf{Target Census Roof Classes}} \\ \cline{2-5}  \\[-1em]
 & \rotatebox{90}{\textbf{Iron Sheets}} & \rotatebox{90}{\begin{tabular}[c]{@{}c@{}}\textbf{Local, Industrial,} \\ \textbf{and Asbestos Tiles}\end{tabular}} & \rotatebox{90}{\textbf{Concrete}} & \rotatebox{90}{\textbf{Grass}} \\ \hline \\[-1em]
\begin{tabular}[c]{@{}l@{}}Rudimentary, basic or \\ unplanned buildings\end{tabular} & \checkmark &  &  & \checkmark \\
\begin{tabular}[c]{@{}l@{}}Building in block structure or\\ large courtyard buildings\end{tabular} & \checkmark &  &  &  \\
Bungalow-type buildings & \checkmark & \checkmark &  &  \\
Villa-type buildings & \checkmark & \checkmark &  &  \\
\begin{tabular}[c]{@{}l@{}}Low to mid-rise\\ multi-unit buildings\end{tabular} & \checkmark & \checkmark &  &  \\
High-rise buildings & \checkmark &  & \checkmark &  \\
Halls & \checkmark &  &  &  \\ \hline
\end{tabular}%
}
\end{table}

\begin{table}[htpb!]
\centering
\caption{Assumed relationship between available height labels ($Y_{C^{h}}$) and discrete target height classes ($C^{h}_{s_i}$).}
\label{appendixtab:buildingheightmapping}
\resizebox{\columnwidth}{!}{%
\begin{tabular}{lcccccc}
\hline \\[-1em]
\multirow{2}{*}{\textbf{\begin{tabular}[c]{@{}l@{}}Continuous Height \\ Labels in meters\end{tabular}}} &
  \multicolumn{6}{c}{\textbf{Discrete Target Height Classes}} \\ \cline{2-7}  
\multicolumn{1}{l}{
   } & \rotatebox{90}{\textbf{H:1}  
   } & \rotatebox{90}{\textbf{H:2}
   } & \rotatebox{90}{\textbf{H:3} 
   } & \rotatebox{90}{\textbf{HBET:3-6}  
   } & \rotatebox{90}{\textbf{HBET:4-7} 
   } & \rotatebox{90}{\textbf{HBET:8+}  
   } \\ \hline \\[-1em]
$\left(0,6\right)$               & \checkmark   &   \checkmark   &     &          &          &         \\ \\[-1em]
$\left[6,9\right)$         &     & \checkmark   &  \checkmark   &    \checkmark      &          &         \\ \\[-1em]
$\left[9,12\right)$          &     &     & \checkmark   & \checkmark        & \checkmark          &         \\ \\[-1em]
$\left[12,21\right)$              &     &     &     & \checkmark        & \checkmark        &         \\ \\[-1em]
$\left[21,24\right)$                 &     &     &     &          & \checkmark        &    \checkmark     \\ \\[-1em]
$\left[24,+\infty\right)$                   &     &     &     &          &          & \checkmark       \\ \hline
\end{tabular}%
}
\end{table}

\begin{table}[htpb!]
\centering
\caption{Assumed relationship between available building type labels ($Y_{C^{h}}$) and discrete target census wall classes ($C^{h}_{s_i}$).}
\label{tab:tableb3}
\resizebox{\columnwidth}{!}{%
\begin{tabular}{lcccccccc}
\hline \\[-1em]
\multirow{2}{*}{\textbf{Building Type Labels}} & \multicolumn{8}{c}{\textbf{Target Census Wall Classes}} \\ \cline{2-9} \\[-1em]
 & \rotatebox{90}{\textbf{Others}} & \rotatebox{90}{\textbf{Burnt bricks}} & \rotatebox{90}{\textbf{Cement blocks}} & \rotatebox{90}{\textbf{Concrete}} & \rotatebox{90}{\textbf{Stones}} & \rotatebox{90}{\textbf{Sun-dried bricks}} & \rotatebox{90}{\textbf{Timber}} & \rotatebox{90}{\textbf{Wood w/ mud}} \\ \hline \\[-1em]
\begin{tabular}[c]{@{}l@{}}Rudimentary, basic or\\ unplanned buildings\end{tabular} & \checkmark & \checkmark & \checkmark & \checkmark & \checkmark & \checkmark & \checkmark & \checkmark \\
\begin{tabular}[c]{@{}l@{}}Building in block structure or\\ large courtyard buildings\end{tabular} & \checkmark & \checkmark & \checkmark & \checkmark &  & \checkmark &  &  \\
Bungalow-type buildings &  & \checkmark & \checkmark & \checkmark &  & \checkmark &  & \checkmark \\
Villa-type buildings &  &  & \checkmark & \checkmark &  & \checkmark &  &  \\
\begin{tabular}[c]{@{}l@{}}Low to mid-rise \\ multi-unit buildings\end{tabular} &  &  & \checkmark & \checkmark &  &  &  &  \\
High-rise buildings &  &  &  & \checkmark &  &  &  &  \\
Halls &  & \checkmark & \checkmark & \checkmark &  & \checkmark &  &  \\ \hline
\end{tabular}%
}
\end{table}

\end{document}